\documentclass[runningheads,a4paper]{llncs} 

\usepackage[table,pdftex,dvipsnames]{xcolor}
\usepackage[cmex10]{amsmath}
\usepackage{url}
\usepackage[caption=false,font=footnotesize]{subfig}
\usepackage{amsmath}
\usepackage{booktabs}
\usepackage{cite}
\usepackage[utf8]{inputenc}
\usepackage{stfloats}
\usepackage{amssymb}
\usepackage[ruled,linesnumbered,lined,boxed,commentsnumbered]{algorithm2e}
\usepackage{tikz}
\usepackage{multirow}
\usepackage{makecell}

\usetikzlibrary{shapes,arrows,positioning,calc}
\usepackage{tikz}

\usepackage{float}

\usepackage{xargs}                      

\usepackage[acronym]{glossaries}
\newacronym{mdp}{MDP}{Markov Decision Process}
\newacronym{mpc}{MPC}{Model Predictive Control}
\newacronym{isp}{IS}{Importance Splitting}
\newacronym{pso}{PSO}{Particle Swarm Optimization}

\newcommand{\xv}{{\boldsymbol x}}
\newcommand{\vv}{{\boldsymbol v}}
\newcommand{\va}{{\boldsymbol {a}}}

\usepackage[colorinlistoftodos,prependcaption,textsize=tiny]{todonotes}
\newcommandx{\unsure}[2][1=]{\todo[linecolor=red,backgroundcolor=red!25,bordercolor=red,#1]{#2}}
\newcommandx{\change}[2][1=]{\todo[linecolor=blue,backgroundcolor=blue!25,bordercolor=blue,#1]{#2}}
\newcommandx{\info}[2][1=]{\todo[linecolor=OliveGreen,backgroundcolor=OliveGreen!25,bordercolor=OliveGreen,#1]{#2}}
\newcommandx{\improvement}[2][1=]{\todo[linecolor=Plum,backgroundcolor=Plum!25,bordercolor=Plum,#1]{#2}}
\newcommandx{\thiswillnotshow}[2][1=]{\todo[disable,#1]{#2}}

\newcommand{\cmmnt}[1]{}

\usepackage{url}
\urldef{\mailsa}\path|{alfred.hofmann, ursula.barth, ingrid.haas, 
frank.holzwarth,|
	\urldef{\mailsb}\path|anna.kramer, leonie.kunz, christine.reiss, 
	nicole.sator,|
	\urldef{\mailsc}\path|erika.siebert-cole, peter.strasser, 
	lncs}@springer.com|    

\DeclareMathOperator*{\argmin}{\arg\min}

\renewcommand{\refname}{\normalfont\selectfont\normalsize\textbf{References}} 

\begin{document}
	\mainmatter 
	\title{ARES: Adaptive Receding-Horizon \\ Synthesis of Optimal Plans}
%
%
	
%
%
%

\author{Anna Lukina\inst{1}	\and Lukas Esterle\inst{1} \and Christian 
Hirsch\inst{1} \and Ezio Bartocci\inst{1} \and \\ Junxing Yang\inst{2} \and Ashish 
Tiwari\inst{3}  \and Scott A. Smolka\inst{2} \and Radu Grosu\inst{1,2}}
\authorrunning{Lukina, Esterle, Hirsch, Bartocci, Yang, Tiwari, Smolka, Grosu}
\titlerunning{ARES: Adaptive Receding-Horizon Synthesis of Optimal Plans}

\institute{Cyber-Physical Systems Group, Technische Universit\"at Wien, Austria 
	\and Department of Computer Science, Stony Brook University, USA \and SRI International, USA
	}
	
	\toctitle{Lecture Notes in Computer Science}
	\tocauthor{Authors' Instructions}
	\maketitle
	
\begin{abstract}

We introduce ARES, an efficient approximation algorithm for 
generating optimal plans (action sequences) that take an 
initial state of a \gls{mdp} to a state 
whose cost is below a specified (convergence) threshold.
ARES uses Particle Swarm Optimization, with \emph{adaptive sizing} for 
both the receding horizon and the particle swarm.  Inspired 
by Importance Splitting, the length of the horizon and the number of 
particles are chosen such that at least one particle reaches a 
\emph{next-level} state, that is, a state where the cost decreases by a 
required delta from the previous-level state.  The level relation on 
states and the plans constructed by ARES implicitly define a Lyapunov 
function and an optimal policy, respectively, both of which could be 
explicitly generated by applying ARES to all states of the \gls{mdp}, up 
to some topological equivalence relation.  We also assess the effectiveness
of ARES by statistically evaluating 
its rate of success in generating optimal plans.  The ARES algorithm 
resulted from our desire to clarify if flying in V-formation is a 
flocking policy that optimizes energy conservation, clear view, and velocity 
alignment.
That is, we were interested to see if one could find optimal plans
that bring a flock from an arbitrary initial state to a
state exhibiting a single connected V-formation.  For flocks with 7 birds, ARES
is able to generate a plan that leads to a
V-formation in 95\% of the 8,000 random initial 
configurations within 63 seconds, on average. 
ARES can also be easily
customized into a model-predictive controller (MPC) with an adaptive 
receding horizon and statistical guarantees of convergence. To the 
best of our knowledge, our adaptive-sizing approach is the first 
to provide \emph{convergence guarantees} in receding-horizon 
techniques. 

\end{abstract}



	\section{Introduction}
\label{sec:intro}

\glsresetall

Flocking or swarming in groups of social animals (birds, fish, ants, bees, etc.) that results in a particular global formation is an emergent collective behavior that continues to fascinate researchers~\cite{Bajec2009AB, Chazelle:2014}.  One would like to know if such a formation serves a higher purpose, and, if so, what that purpose is.

One well-studied flight-formation behavior is \emph{V-formation}.  Most of the work in this area has concentrated on devising simple dynamical rules that, when followed by each bird, eventually stabilize the flock to the desired V-formation~\cite{flake1998computational,dimock2003aerodynamic,nathan2008}. This approach, however, does not shed very much light on the overall purpose of this emergent behavior.


In previous work~\cite{yang2016bda,yang2016love}, we hypothesized that flying in V-formation is nothing but an optimal policy for a flocking-based Markov Decision Process (MDP) $\mathcal{M}$.  States of $\mathcal{M}$, at discrete time $t$, are of the form $(\xv_i(t),\vv_i(t))$, $1\,{\leqslant}\,i \,{\leqslant}\,N$, where $\xv_i(t)$ and $\vv_i(t)$ are $N$-vectors (for an $N$-bird flock) of 2-dimensional positions and velocities, respectively. $\mathcal{M}$'s transition relation, shown here for bird~$i$ is simply and generically given by
\vspace*{-1mm}\begin{eqnarray*}
\label{eq:model}
 \xv_i(t + 1) &=& \xv_i(t) + \vv_i(t+1)\label{eq:x},\\
 \vv_i(t + 1) &=& \vv_i(t) + \va_i(t)\label{eq:v},\\[-6mm]
\end{eqnarray*}
where $\va_i(t)$ is an action, a 2-dimensional acceleration in this case, that bird~$i$ can take at time $t$.  $\mathcal{M}$'s cost function reflects the energy-conservation, velocity-alignment and clear-view benefits enjoyed by a state of $\mathcal{M}$ (see Section~\ref{sec:vform}).

In this paper, we not only confirm this hypothesis, but we also devise a very general \emph{adaptive, receding-horizon synthesis algorithm} (ARES) that, given an MDP and one of its initial states, generates an optimal plan (action sequence) taking that state to a state whose cost is below a desired threshold.  In fact, ARES implicitly defines an \emph{optimal, online-policy, synthesis algorithm} that could be used in practice if plan generation can be performed in real-time.

ARES makes repeated use of \gls{pso}~\cite{Kennedy95particleswarm} to effectively generate a plan. This was in principle unnecessary, as one could generate an optimal plan by calling \gls{pso} only once, with a maximum plan-length horizon.  Such an approach, however, is in most cases impractical, as every unfolding of the MDP adds a number of new dimensions to the search space.  Consequently, to obtain an adequate coverage of this space, one needs a very large number of particles, a number that is either going to exhaust available memory or require a prohibitive amount of time to find an optimal plan.

\begin{figure}[t]
	\centering
	\usetikzlibrary{calc}
\begin{tikzpicture}[scale=0.75, xscale=0.9, dot/.style={draw,circle,minimum size=1mm,inner sep=0pt,outer sep=0pt,fill,gray}]
\tikzset{
  cell/.style = {label={[font=\small]}}
}
\draw (0,-0.75);
\draw[-latex] (-0.85,0) -- (2*pi+0.75,0) node[below left] {\scriptsize$State$};
\draw[-latex] (-0.75,-0.1) -- (-0.75,6) node[left] {\footnotesize{\scriptsize$Cost$}};

\draw [red, thick, domain=-0.6:(7*pi/4-0.95), samples=100] plot (\x, {3.5+1.1*cos((\x-2) r) * (\x-2)});

\foreach \x [count=\k from 1] in {7*pi/4-0.95, 4.8, 5.0, 6.5}
{
	\coordinate (x\k) at (\x,{3.5+1.1*cos((\x-2) r) * (\x-2)});
}

\coordinate (xnm1) at (5.8,0.2);
\coordinate (xn) at (6.5,0.15);

\draw [red, thick] plot [smooth,tension=0.75] coordinates{
	(x1)
	($(x2) ! 0.1 ! (xn) $)
	(xnm1)
	(xn)
};

\foreach \x [count=\k from 1] in {-0.4, 0.11, pi-2, pi-2+1.06, pi-2+2.12, pi-2+3.2}
{
	\coordinate [dot] (L\k) at (\x,{3.5+1.1*cos((\x-2) r) * (\x-2)});
	\draw [gray,-] (\x,0) -- (L\k);
	\coordinate (Sk\k) at (\x,0);
}



\node [cell, anchor=north] at (Sk1) {\scriptsize$s_{0}$};
\node (s1) [cell, anchor=north] at (Sk2) {\scriptsize$s_{1}$};

\draw (L1) ++(-0.2,0) node [anchor=east] {\scriptsize$\ell_{0}$};
\draw (L2) ++(0.15,0) node [anchor=east] {\scriptsize$\ell_{1}$};

\node (si) [anchor=north] at (Sk3) {\scriptsize$s_{i}$};
\node (si3) [anchor=north] at (Sk6) {\scriptsize$s_{i+3}$};
\node at ($ (s1) ! 0.5 ! (si) $) {\scriptsize\dots};
\node at ($ (si) ! 0.5 ! (si3) $) {\scriptsize\dots};

\node [anchor=north east] at (L3) {\scriptsize$\ell_{i}$};
\node [anchor=east] at (L6) {\scriptsize$\ell_{i+1}$};

\end{tikzpicture}
    \begin{tikzpicture}[scale=0.75, dot/.style={draw,circle,minimum size=1mm,inner sep=0pt,outer sep=0pt, fill, gray}]
\draw (0,-0.75);
\draw[-latex] (-0.1,0) -- (5,0) node[below left] {\scriptsize$State$};
\draw[-latex] (0,-0.1) -- (0,6) node[left] {\scriptsize$Level$};

\draw [red, thick, domain=0:5] plot (\x, {2*exp(1-\x)});

\foreach \x [count=\k from 0] in {0.0, 0.272, 0.635, 1.16, 3}
{
	\coordinate [dot] (e\k) at (\x, {2*exp(1-\x)});
	\coordinate (i\k) at (\x, 0);
	\coordinate (ilabel\k) at (\x, -0.1);

	\draw [-, gray] (0, {2*exp(1-\x)}) -- (e\k);
};

\node [below] at (ilabel0) {\scriptsize $s_{0}$};
\node [below] at (ilabel1) {\scriptsize $s_{1}$};
\node [below] at (ilabel2) {\scriptsize $s_{i}$};
\node [below] at (ilabel3) {\scriptsize $s_{i+3}$};

\node (ilabelk) [below] at (3, -0.15) {\scriptsize $s^{*}$};
\node [below] at ($ (ilabel3) ! 0.5 ! (ilabelk) $) {\scriptsize$\dots$};

\foreach \x [count=\k from 0] in {0.0, 0.272, 0.635, 1.16}
{
	\coordinate (llabel\k) at (0, {2*exp(1-\x)});
};

\node [anchor=east] at (llabel0) {\scriptsize{$\ell_{0}$}};
\node [anchor=east] at (llabel1) {\scriptsize{$\ell_{1}$}};
\node [anchor=east] at (llabel2) {\scriptsize{$\ell_{i}$}};
\node [anchor=east] at (llabel3) {\scriptsize{$\ell_{i+1}$}};

\node [anchor=east] at ($ (llabel1) ! 0.375 ! (llabel2) $) {\scriptsize$\vdots$};


\draw [-, gray] (e1) -- (i1);
\draw [-, gray] (e2) -- (i2);
\draw [-, gray] (e3) -- (i3);



\draw [-] (3,-0.1) -- (3,{2*exp(1-3)+0.1}) node [anchor=south west, pos=0.5] {\scriptsize$\varphi$};
\draw [-latex] (3,0) ++(270:0.1) -- (3,0);
\draw [-latex] (3,{2*exp(1-3)}) ++(90:0.1) -- (3,{2*exp(1-3)});
\coordinate (llabelk) at (0, {exp(1-3)});
\node [anchor=east] at (llabelk) {\scriptsize{$\ell_m$}};

\node [anchor=east] at ($ (llabel3) ! 0.375 ! (llabelk) $) {\scriptsize$\vdots$};

\end{tikzpicture}
   \vspace*{-3mm}
\caption{Left: If state $s_0$ has cost $\ell_0$, and its successor-state $s_{1}$ has cost less than $\ell_{1}$, then a horizon of length~1 is appropriate. However, if $s_i$ has a local-minimum cost $\ell_i$, one has to pass over the cost ridge in order to reach level $\ell_{i+1}$, and therefore ARES has to adaptively increase the horizon to~3. Right: The cost of the initial state defines $\ell_0$ and the given threshold $\varphi$ defines $\ell_{m}$. By choosing $m$ equal segments on an asympthotically converging (Lyapunov) function (where the number $m$ is empirically determined), one obtains on the vertical cost-axis the levels required for ARES to converge.}   
	\label{fig:levels}
    \vspace*{-5mm}
\end{figure}
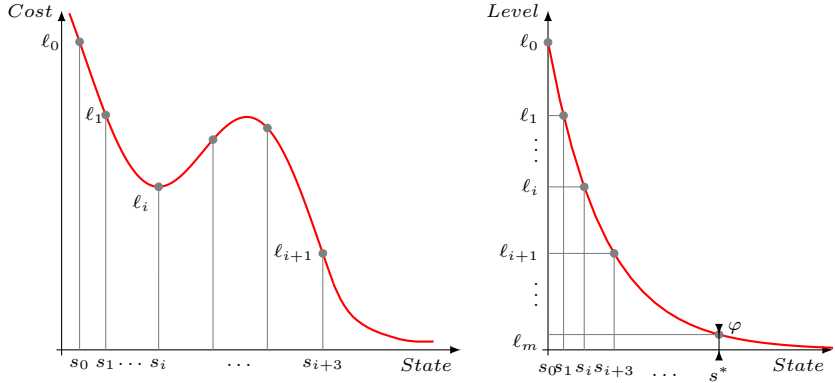

A simple solution to this problem would be to use a short horizon, typically of size two or three.  This is indeed the current practice in \gls{mpc}~\cite{mpc1989}. This approach, however, has at least three major drawbacks.  First, and most importantly, it does not guarantee convergence and optimality, as one may oscillate or become stuck in a local optimum.  Second, in some of the steps, the window size is unnecessarily large thereby negatively impacting performance. Third, in other steps, the window size may be not large enough to guide the optimizer out of a local minimum (see Fig.~\ref{fig:levels}~(left)). One would therefore like to find the proper window size adaptively, but the question is how one can do it.

Inspired by \gls{isp}, a sequential Monte-Carlo technique for estimating the probability of rare events, we introduce the notion of a \emph{level-based horizon} (see Fig.~\ref{fig:levels}~(right)). Level $\ell_0$ is the cost of the initial state, and level $\ell_m$ is the desired threshold.  By using a state function, asymptotically converging to the desired threshold, we can determine a sequence of levels, ensuring convergence of ARES towards the desired optimal state(s) having a cost below $\ell_m\,{=}\,\varphi$.
%

The levels serve two purposes.  First, they implicitly define a Lyapunov function, which guarantees convergence.  If desired, this function can be explicitly generated for all states, up to some topological equivalence.  Second, the levels help PSO overcome local minima (see Fig.~\ref{fig:levels}~(left)).  If reaching a next level requires PSO to temporarily pass over a state-cost ridge, ARES incrementally increases the size of the horizon, up to a maximum length.

Another idea imported from \gls{isp} is to maintain $n$ clones of the initial state at a time, and run \gls{pso} on each of them (see Fig.~\ref{fig:approach}). This allows us to call \gls{pso} for each clone and desired horizon, with a very small number of particles per clone.  Clones that do not reach the next level are discarded, and the successful ones are resampled.  The number of particles is increased if no clone reaches a next level, for all horizons chosen.  Once this happens, we reset the horizon to one, and repeat the process.  In this way, we adaptively focus our resources on escaping from local minima.  At the last level, we choose the optimal particle (a V-formation in case of flocking) and traverse its predecessors to find a plan.
 
We asses the rate of success in generating optimal plans in form of an $(\varepsilon,\delta)$-approximation scheme, for a desired error margin $\varepsilon$, and confidence ratio $1{-}\delta$.  Moreover, we can use the state-action pairs generated during the assessment (and possibly some additional new plans) to construct an explicit (tabled) optimal policy, modulo some topological equivalence.  Given enough memory, one can use this policy in real time, as it only requires a table look-up.

To experimentally validate our approach, we have applied ARES to the problem of V-formation in bird flocking (with a deterministic MDP).  The cost function to be optimized is defined as a weighted sum of the (flock-wide) clear-view, velocity-alignment, and upwash-benefit metrics. Clear view and velocity alignment are more or less obvious goals. Upwash optimizes energy savings.  By flapping its wings, a bird generates a trailing upwash region off its wing tips; by using this upwash, a bird flying in this region (left or right) can save energy. Note that by requiring that at most one bird does not feel its effect, upwash can be used to define an analog version of a connected graph.

We ran ARES on 8,000 initial
states chosen uniformly and at random, such that they are packed closely enough to feel upwash, but not too close to collide.  We succeeded to generate a V-formation 95\% of the time, with an error margin of~0.05 and a confidence ratio of~0.99.  These error margin and confidence ratio dramatically improve if we consider all generated states and the fact that each state within a plan is independent from the states in all other plans.

The rest of this paper is organized as follows. Section~\ref{sec:vform} reviews our work on bird flocking and V-formation, and defines the manner in which we measure the cost of a flock (formation).  Section~\ref{sec:swarmOptimization} revisits the swarm optimization algorithm used in this paper, and Section~\ref{sec:importanceSplitting} examines the main characteristics of importance splitting. Section~\ref{sec:problem} states the definition of the problem we are trying to solve. Section~\ref{sec:ares} introduces ARES, our adaptive receding-horizon synthesis algorithm for optimal plans, and discusses how we can extend this algorithm to explicitly generate policies. Section~\ref{sec:results} measures the efficiency of ARES in terms of an $(\varepsilon,\delta)$-approximation scheme. Section~\ref{sec:related} compares our algorithm to related work, and Section~\ref{sec:conclusion} draws our conclusions and  discusses future work.
               
	\newcommand{\VM}{{\it VM}}
\newcommand{\CV}{{\it CV}}
\newcommand{\UB}{{\it UB}}
\section{V-Formation MDP}
\label{sec:vform}
We represent a flock of birds as a dynamically evolving system. Every
bird in our model~\cite{grosu2014isola} moves in 2-dimensional space performing
acceleration actions determined by a global controller. Let $\xv_i(t),
\vv_i(t)$ and $\va_i(t)$ be 2-dimensional vectors of  positions, velocities, and accelerations, respectively, of bird $i$ at time $t$, where $i\,{\in}\,\{1,\ldots,b\}$, 
for a fixed $b$.  The discrete-time behavior of bird $i$ is then
\begin{align}
\xv_i(t + 1) &= \xv_i(t) + \vv_i(t + 1),\notag\\
\vv_i(t + 1) &= \vv_i (t)+ \va_i(t).
\end{align}

\vspace*{-1mm}\noindent{}The controller detects the positions and velocities 
of all birds through sensors, and uses this information to compute an optimal 
acceleration for the entire flock.  A bird uses its own component of the
solution to update its velocity and position.

\begin{figure}[t]
\centering	
\includegraphics[width=.28\textwidth]{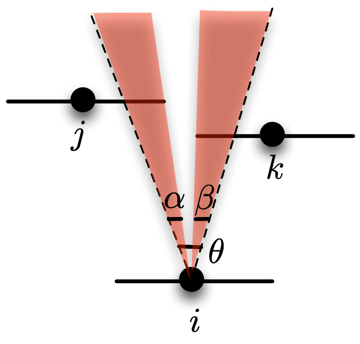}
\includegraphics[width=.3\textwidth]{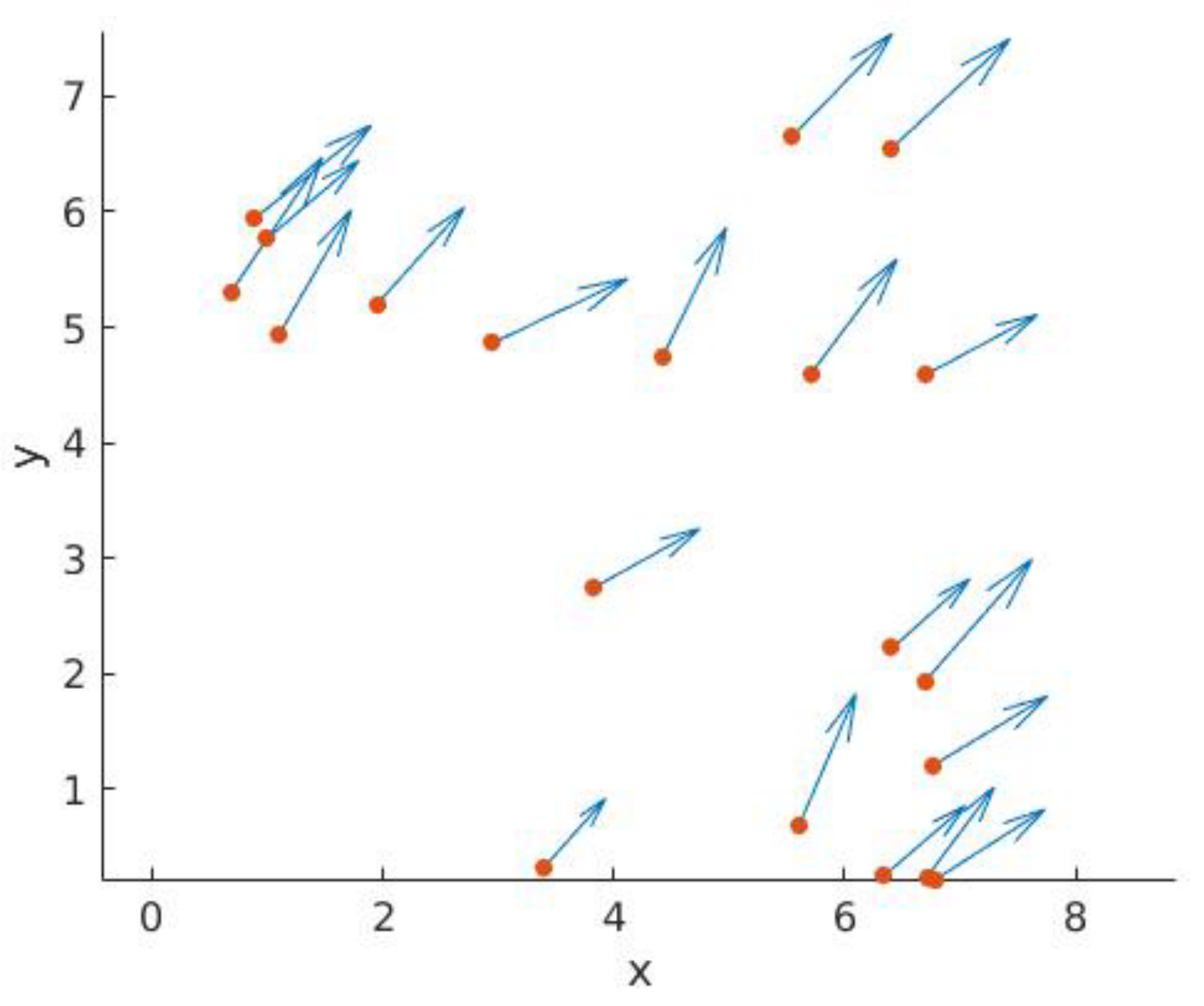}
\includegraphics[width=.36\textwidth]{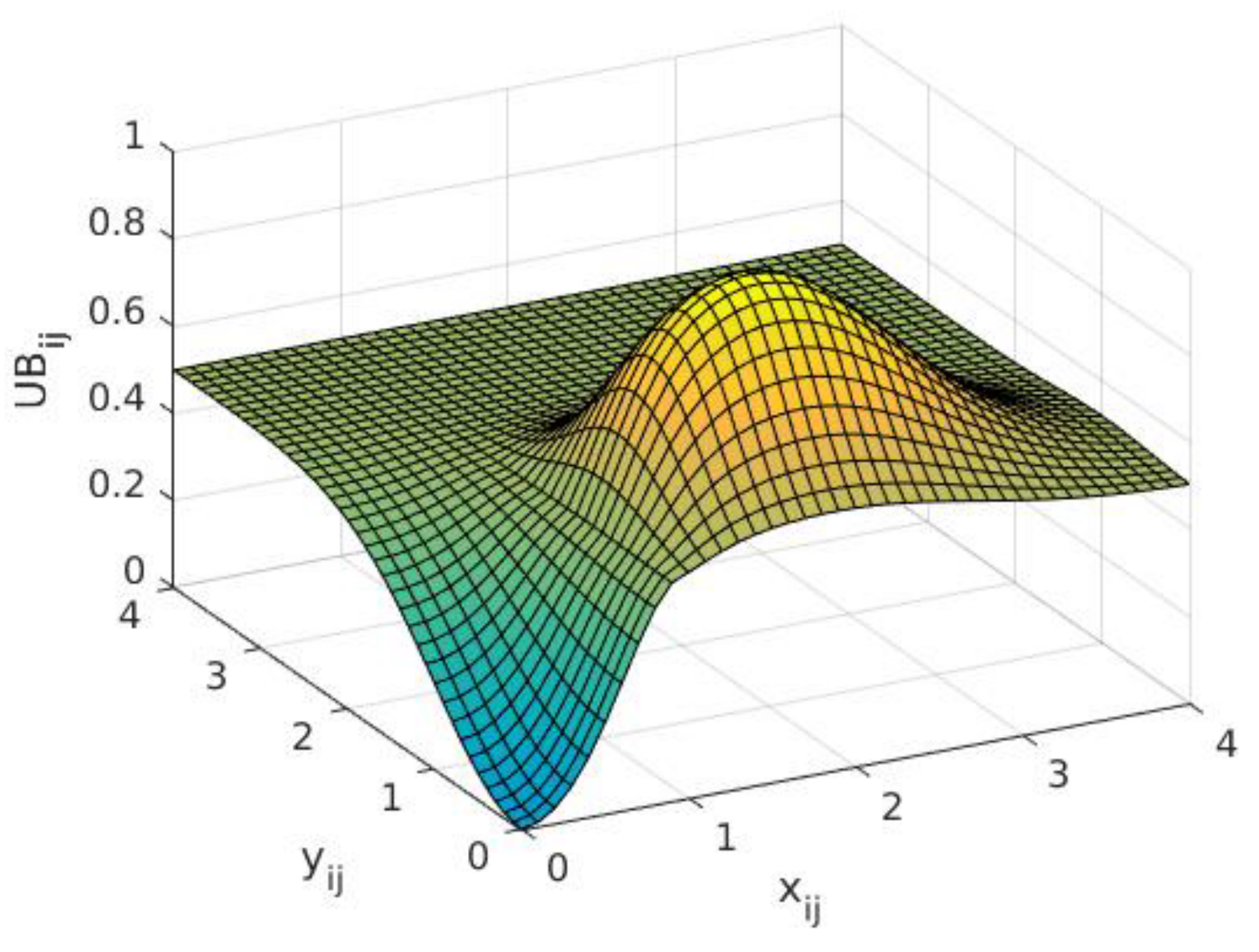}
\vspace*{-1mm}
\caption{
Illustration of the clear view ($\CV$), velocity matching  
($\VM$), and upwash benefit ($\UB$) metrics. 
Left: Bird $i$'s view is partially blocked by birds $j$ 
and $k$. Hence, its clear view is $\CV\,{=}\,(\alpha\,{+}\,\beta)/\theta$. 
Middle: A flock and its unaligned bird velocities results in a 
velocity-matching metric $\VM\,{=}\,6.2805$.  In contrast, $\VM\,{=}\,0$
when the velocities of all birds are aligned. 
Right: Illustration of the (right-wing) upwash benefit bird $i$ receives from bird $j$ depending on how it is positioned behind bird $j$.  Note that  
bird $j$'s downwash region is directly behind it.
}
\label{fig:fitness}
\vspace*{-4mm}
\end{figure}

We extend this discrete-time dynamical model to a (deterministic) MDP by adding a
cost (fitness) function\footnote{A classic 
MDP~\cite{russellnorvig} is obtained by adding sensor/actuator or wind-gust noise.} based on the following metrics inspired by~\cite{yang2016love}:

\begin{itemize}
	\vspace*{-1.5mm}\item \emph{Clear View} ($\CV$). A bird's visual field is a cone with 
    angle $\theta$ that can be blocked by the wings of other birds.  We define
    the clear-view metric by accumulating the percentage of a bird's visual 
    field that is blocked by other birds.  Fig.~\ref{fig:fitness} (left) illustrates 
    the calculation of the clear-view metric.  The optimal value in a V-formation 
    is $\CV^*{=}\,0$, as all birds have a clear view.
	\vspace*{1mm}\item \emph{Velocity Matching} ($\VM$). The accumulated 
	differences between the velocity of each bird and all other birds, 
    summed up over all birds in the flock defines $\VM$. Fig.~\ref{fig:fitness} (middle) depicts the values of $\VM$ in a velocity-unmatched flock. The optimal
    value in a V-formation is $\VM^*{=}\,0$, as all birds will have the same 
    velocity (thus maintaining the V-formation).
	\vspace*{1mm}\item \emph{Upwash Benefit} ($\UB$). The trailing upwash is 
    generated near the wingtips of a bird, while downwash is generated near 
    the center of a bird.  We accumulate all birds' upwash benefits using a 
    Gaussian-like model of the upwash and downwash region, as shown in 
    Fig.~\ref{fig:fitness} (right) for the right wing. The maximum upwash a 
    bird can obtain has an upper bound of~1. For bird~$i$ with $\UB_i$, we 
    use $1\,{-}\UB_i$ as its upwash-benefit metric, because the optimization 
    algorithm performs minimization of the fitness metrics. The optimal value 
    in a V-formation is $\UB^*{=}\,1$, as the leader does not receive any upwash.
\end{itemize}

\vspace*{-1.5mm}\noindent{}Finding smooth and continuous formulations 
of the fitness metrics is a key element of solving optimization problems.  The PSO 
algorithm has a very low probability of finding an optimal solution if 
the fitness metric is not well-designed.

Let $\boldsymbol{c}(t)=\{\boldsymbol{c}_i(t)\}_{i=1}^b=\{\xv_i(t), \vv_i(t)\}_{i=1}^b$ be a flock 
configuration at time-step $t$. Given the above metrics, the overall fitness (cost) metric $J$ 
is of a sum-of-squares combination of $\VM$, $\CV$, and $\UB$ defined as follows:
\vspace*{-1mm}
\begin{align}
J(\boldsymbol{c}(t),\va^h(t),{h}) = (\CV(\boldsymbol{c}_{\va}^{h}(t))-\CV^*)^2 &+ 
(\VM(\boldsymbol{c}_{\va}^{h}(t))-\VM^*)^2 \nonumber \\ & +(\UB(\boldsymbol{c}_{\va}^{h}(t))-\UB^*)^2,
\label{eq:fitness}
\end{align}
\noindent{}where ${h}$ is the receding prediction horizon (RPH), $\va^h(t)$ is a sequence of accelerations of length ${h}$, and $\boldsymbol{c}_{\va}^{h}(t)$ is 
the configuration reached after applying $\va^h(t)$ to $\boldsymbol{c}(t)$.
Formally, we have\vspace*{-2mm}
\begin{align}
\boldsymbol{c}_{\va}^{h}(t)= \{\xv_{\va}^{h}(t), \vv_{\va}^{h}(t)\} = \{\xv(t)+\sum_{\tau=1}^{{h}(t)}\vv(t+\tau), \vv(t)+\sum_{\tau=1}^{{h}(t)} \va^\tau(t) \},
\end{align}
where $\va^\tau(t)$ is the $\tau$th acceleration of $\va^h(t)$.
A novelty of this paper is that, as described in Section~\ref{sec:ares}, we allow  RPH ${h}(t)$ to be \emph{adaptive} in nature. 

The fitness function $J$ has an optimal value of $0$ in a perfect V-formation. The main goal of ARES is to compute the sequence of acceleration actions that lead the flock from a random initial configuration towards 
a controlled V-formation characterized by optimal fitness in order to conserve energy during flight including 
optimal combination of a clear visual field along with visibility of lateral
neighbors.  Similar to the centralized version of the approach given in~\cite{yang2016love}, ARES performs a single flock-wide minimization of $J$ at each time-step $t$ to obtain an optimal plan of  length $h$ of acceleration actions:\vspace*{-3mm}

\begin{align}
&\textbf{opt-$\va$}^{h}(t)=\{\textbf{opt-$\va$}_i^{h}(t)\}_{i=1}^{b}=\argmin_{\va^h(t)}J(\boldsymbol{c}(t),\va^h(t),{h}).
\label{eq:opt}
\end{align}
\vspace*{-3mm}\noindent{} 

The optimization is subject to the following constraints on the maximum 
velocities and accelerations: $||\vv_i(t)||\,{\leqslant}\,\vv_{max},
||\va^h_i(t)||\,{\leqslant}\,\rho||\vv_i(t)||$ $\forall$ $i\,{\in}\,\{1,\ldots,b\}$,
where $\vv_{max}$ is a constant and $\rho\,{\in}\,(0,1)$. 
The initial positions and velocities of each bird are selected at random 
within certain ranges, and limited such that the distance between any 
two birds is greater than a (collision) constant $d_{min}$, and small
enough for all birds, except for at most one, to feel the $\UB$. 
In the following sections, we demonstrate how to generate optimal plans taking the 
initial state to a stable state with optimal fitness. 
         
    \section{Particle Swarm Optimization}
\label{sec:swarmOptimization}

Particle Swarm Optimization (PSO) is a randomized approximation algorithm for computing the value of a parameter minimizing a possibly nonlinear cost (fitness) function.  Interestingly, PSO itself is inspired by bird flocking~\cite{Kennedy95particleswarm}.  Hence, PSO assumes that it works with a flock of birds. 

Note, however, that in our running example, these birds are ``acceleration birds'' (or particles), and not the actual birds in the flock. Each bird has the same goal, finding food (reward), but none of them knows the location of the food. However, every bird knows the distance (horizon) to the food location. PSO works
by moving each bird preferentially toward the bird closest to food.

ARES uses Matlab-Toolbox $\texttt{particleswarm}$, which performs the classical version of PSO. This PSO creates a swarm of particles, of size say ${p}$, uniformly at random within a given bound on their positions and velocities. Note that in our example, each particle represents itself a flock of bird-acceleration sequences $\{\va_i^{{h}}\}_{i=1}^b$, where ${h}$ is the current length of the receding horizon. PSO further chooses a neighborhood of a random size for each particle $j$, $j\,{=}\,\{1,\ldots,p\}$, and computes the fitness of each particle. Based on the fitness values, PSO stores two vectors for $j$: its so-far personal-best position $\mathbf{x}_{P}^j(t)$, and its fittest neighbor's position $\mathbf{x}_{G}^j(t)$. The positions and velocities of each particle $j$ in the particle swarm $1\,{\leqslant}\,j\,{\leqslant}\,p$ are updated according to the following rule:

\vspace*{-4mm}
\begin{align}
\mathbf{v}^j(t+1) = \omega\cdot\mathbf{v}^j(t) &+ y_1\cdot \mathbf{u_1}(t+1)\otimes(\mathbf{x}_{P}^j(t)-\mathbf{x}^j(t))  \nonumber \\
&+ y_2\cdot \mathbf{u_2}(t+1)\otimes(\mathbf{x}_{G}^j(t)-\mathbf{x}^j(t)),
\label{eq:swarm}
\end{align}

\vspace*{-1mm}\noindent{}where $\omega$ is \emph{inertia weight}, which determines the trade-off between global and local exploration of the swarm (the value of $\omega$ is proportional to the exploration range); $y_1$ and $y_2$ are \emph{self adjustment} and \emph{social adjustment}, respectively; $\mathbf{u_1},\mathbf{u_2}\,{\in}\,{\rm Uniform}(0,1)$ are randomization factors; and $\otimes$ is the vector dot product, that is, $\forall$ random vector $\mathbf{z}$: $(\mathbf{z}_1,\ldots,\mathbf{z}_b)\otimes(\mathbf{x}_1^j,\ldots,\mathbf{x}_b^j)=(\mathbf{z}_1\mathbf{x}_1^j,\ldots,\mathbf{z}_b\mathbf{x}_b^j)$. 

If the fitness value 
for $\mathbf{x}^j(t+1)\,{=}\,\mathbf{x}^j(t)\,{+}\,\mathbf{v}^j(t+1)$ is lower than the one for $\mathbf{x}_{P}^j(t)$, then $\mathbf{x}^j(t+1)$ is assigned to $\mathbf{x}_{P}^j(t+1)$. The particle with the best fitness over the whole swarm becomes a global best for the next iteration. The procedure is repeated until the number of iterations reaches its maximum, the time elapses, or the minimum criteria is satisfied. For our bird-flock example we obtain in this way the best acceleration.

    \section{Importance Splitting}
\label{sec:importanceSplitting}

Importance Splitting (\gls{isp}) is a sequential Monte-Carlo approximation technique for estimating the probability of rare events 
in a Markov process%
~\cite{CerouGuyader2007}.
The algorithm uses a sequence 
$S_0, S_1, S_2, \ldots, S_m$ of sets of states (of increasing ``importance'') such that $S_0$ is the set of initial states and $S_m$ is the set of states defining the rare event.
The probability $p$, computed as $\mathbf{P}(S_m\,|\,S_0)$ of reaching $S_m$ from the initial set of states $S_0$, is assumed to be extremely low (thus, a rare event), and one desires to estimate this probability~\cite{Glasserman1999}. Random sampling approaches, such as the additive-error approximation algorithm described in Section~\ref{sec:results}, are bound to fail (are intractable) in this case, as they would require an enormous number of samples to estimate $p$ with low-variance. 

Importance splitting is a way of decomposing the estimation of $p$.
In IS,
the sequence $S_0,S_1,\ldots$ of sets of states is defined so that the conditional probabilities $p_i\,{=}\,\mathbf{P}(S_i\,|\,S_{i-1})$
of going from one level, $S_{i-1}$, to the next one, $S_i$, are considerably larger than $p$, and essentially equal to one another.  The resulting probability of the rare event is then calculated as the product  $p\,{=}\,\prod_{i=1}^{k}p_i$ of the intermediate probabilities. The levels can be defined adaptively~\cite{KalajdzicIsola16}. 

To estimate $p_i$, IS uses a swarm of particles of size $N$, with a given initial distribution over the states of the stochastic process.  During stage $i$ of the algorithm, each particle starts at level $S_{i-1}$ 
and traverses the states of the stochastic process, 
checking if it reaches $S_i$.
If, at the end of the stage, the particle fails to reach $S_i$,
the particle is discarded. Suppose that $K_i$ particles survive. In this case, $p_i\,{=}\,K_i{/}N$. Before starting the next stage,
the surviving particles are resampled, such that IS once again has $N$ particles.  
Whereas IS is used for estimating probability of a rare event in a Markov process, we use it here for synthesizing a plan for a {\em{controllable}} Markov process, by combining it with ideas from controller synthesis (receding-horizon control) and nonlinear optimization (PSO). 

    \section{Problem Definition}
\label{sec:problem}
\begin{definition}A \textbf{Markov decision process (MDP)} $\mathcal{M}$ is a sequential decision problem that consists of a set of states $S$ (with an initial state $s_0$), a set of actions $A$, a transition model $T$, and a cost function $J$. An MDP is \textbf{deterministic} if for each state and action, $T\,{:}\:S\,{\times}\,{A}\,{\rightarrow}\,{S}$ specifies a unique state.
\end{definition}


\begin{definition}
The \textbf{optimal plan synthesis problem} for an MDP $\mathcal{M}$, an arbitrary initial state $s_0$ of $\mathcal{M}$, and a threshold $\varphi$ is to synthesize a sequence of actions $\va^{i}$ of length $1\,{\leqslant}\,i\,{\leqslant}\,m$ taking $s_0$ to a state $s^{*}$ such that cost $J(s^{*})\,{\leqslant}\,\varphi$.  
\end{definition}

Section~\ref{sec:ares} presents our adaptive receding-horizon synthesis algorithm (ARES)
for the optimal plan synthesis problem.
In our flocking example (Section~\ref{sec:vform}), ARES is used to synthesize a sequence of acceleration-actions bringing an arbitrary bird flock $s_0$ to an optimal state of V-formation $s^{*}$. We assume that we can easily extend such an optimal plan to maintain the cost of successor states below $\varphi$ ad infinitum (optimal stability).
    \section{The ARES Algorithm for Plan Synthesis}
\label{sec:ares}

As mentioned in Section~\ref{sec:intro}, one could in principle solve the optimization problem defined in Section~\ref{sec:problem} by calling the PSO only once, with a horizon $h$ in $\mathcal{M}$ equaling the maximum length $m$ allowed for a plan. This approach, however, tends to explode the search space, and is therefore in most cases intractable. Indeed, preliminary experiments with this technique applied to our running example could not generate any convergent plan.  

A more tractable approach is to make repeated calls to PSO with a small horizon length $h$.  The question is how small $h$ can be. \emph{The current practice in model-predictive control (MPC) is to use a fixed} $h$, $1\,{\leqslant}\,h\,{\leqslant}\,3$ (see the outer loop of Fig.~\ref{fig:approach}, where resampling and conditional branches are disregarded). Unfortunately, this forces the selection of \emph{locally-optimal plans} (of size less than three) in each call, and there is \emph{no guarantee of convergence} when joining them together.  In fact, in our running example, we were able to find plans leading to a V-formation in only $45\%$ of the time for $10,000$ random initial flocks.

Inspired by \gls{isp} (see Fig.~\ref{fig:levels}~(right) and Fig.~\ref{fig:approach}), we introduce the notion of a \emph{level-based horizon}, where level $\ell_0$ equals the cost of the initial state, and level $\ell_m$ equals the threshold $\varphi$.  Intuitively, by using an asymptotic cost-convergence function ranging from $\ell_0$ to $\ell_{m}$, and dividing its graph in $m$ equal segments, we can determine on the vertical axis a sequence of levels ensuring convergence. 

The asymptotic function ARES implements is essentially $\ell_{i}\,{=}\,\ell_0\,(m-i){/}\,m$, but specifically tuned for each particle. Formally, if particle $k$ has previously reached level equaling $J_k(s_{i-1})$, then its next target level is within the distance $\Delta_k\,{=}\,J_k(s_{i-1}){/}(m\,{-}\,i\,{+}\,1)$. In Fig.~\ref{fig:approach}, after passing the thresholds assigned to them, values of the cost function in the current state $s_i$ are sorted in ascending order $\{\widehat{J}_{k}\}_{k=1}^n$. The lowest cost $\widehat{J}_1$ should be apart from the previous level $\ell_{i-1}$ at least on its $\Delta_1$  for the algorithm to proceed to the next level $\ell_i\,{:=}\,\widehat{J}_1$.
\begin{figure}[t]
\definecolor{ShineSky}{rgb}{0, 0.9, 1}
\definecolor{ShineGrass}{rgb}{0.85, 0.88, 0.59}
\definecolor{InfosysDarkGrey}{gray}{0.4}
\definecolor{InfosysLightGrey}{gray}{0.6}
\definecolor{TuWienBlue}{cmyk}{1,0.38,0,0.15}
\definecolor{TuInfRed}{cmyk}{0,1,1,0}

\tikzstyle{block} = [draw, fill=ShineSky!20, rectangle, 
minimum height=3em, minimum width=6em]
\tikzstyle{conf} = [draw, fill=ShineSky!20, circle, node distance=0.7cm]
\tikzstyle{inv} = [draw, circle, node distance=0.7cm]
\tikzstyle{dots} = [->,>=stealth',shorten >=1pt,auto,node distance=2cm,
main node/.style={thick,circle,draw,font=\sffamily\Large}]
\tikzstyle{process} = [rectangle, minimum width=3cm, minimum height=1cm, text 
centered, draw=black, fill=orange!30]
\tikzstyle{arrow} = [thick,->,>=stealth]
\tikzstyle{pso} = [rectangle, rounded corners, minimum width=0.5cm, minimum 
height=0.5cm,text centered, draw=black, fill=gray!30]
\tikzstyle{fit} = [rectangle, rounded corners, minimum width=0.5cm, minimum 
height=0.5cm,text centered, draw=black, fill=gray!30]
\tikzstyle{box} = [fill=lightgray!45,rounded corners,minimum 
width=1.1cm,minimum height=4.02cm, anchor=north]
\tikzstyle{decision} = [diamond, aspect=1, minimum width=1cm, text 
centered, draw=black, fill=ShineGrass]
\tikzstyle{output} = [coordinate]

\begin{tikzpicture}[auto, node distance=2cm,>=latex', cross/.style={path 
	picture={ 
		\draw[black]
		(path picture bounding box.south east) -- (path picture bounding 
		box.north west) (path picture bounding box.south west) -- (path picture 
		bounding box.north east);
	}}]
	
	\node (start) [conf] {};
    \node [above = .2mm of start,text width=.3cm,align = center]{\scriptsize{$s_0$}};
	
	\node (c1) [conf, right = 5mm of start] {};
	\node (c2) [conf, above of=c1] {};
	\node (c3) [conf, above of=c2] {};
	\node (cn) [conf, below = 15mm of c1] {};
	
	\path (c1) -- node (dots) [auto=false]{\vdots} (cn);
	
	\draw [->] (start) -- (c1);
	\draw [->] ($ (start) !0.5! (c1) $) |- (c2);
	\draw [->] ($ (start) !0.5! (c1) $) |- (c3);
	\draw [->] ($ (start) !0.5! (c1) $) |- (c3);
	\draw [->] ($ (start) !0.5! (c1) $) |- (cn);
	\node [above = 0.2mm of c3,text width=2cm,align = 
	center]{\scriptsize{$s_0$}};
	
	\node (psoBox) [box] at (1.68,1.8) {};
	\node [above = 3mm of c3,text width=2cm,align = center]{\scriptsize{$n$ 
	clones}}; 
	
	\node(pso1) [pso, right=3mm of c1] {\scriptsize{PSO}};
	\node(pso2) [pso, right=3mm of c2] {\scriptsize{PSO}};
	\node(pso3) [pso, right=3mm of c3] {\scriptsize{PSO}};	
	\node(pson) [pso, right=3mm of cn] {\scriptsize{PSO}};
	
	\path (pso1) -- node (dots2) [auto=false]{\vdots} (pson);
	
	\draw [->] (c2) -- (pso2);
	\draw [->] (c3) -- (pso3);
	\draw [->] (c1) -- (pso1);
	\draw [->] (cn) -- (pson);
	
	\node (c1n) [conf, right = 5mm of pso1] {};
	\node (c2n) [conf, above of=c1n] {};
	\node (c3n) [conf, above of=c2n] {};
	\node (cnn) [conf, below = 15mm of c1n] {};
	\path (c1n) -- node (dots3) [auto=false]{\vdots} (cnn); 
	\node [above = 0.1mm of c3n,text width=0.3cm,align = center]	
	{\scriptsize{$s_i$}};
    \node [above = 3mm of c3n,text width=3cm,align = center]{\scriptsize{next states after applying $\{\va^h_k\}_{k=1}^n$}};
	
	
	\draw [->, above] (pso2) -- node {\scriptsize{$\va_{2}^{h}$}} (c2n);
	\draw [->] (pso3) -- node {\scriptsize{$\va_{1}^{h}$}} (c3n);
	\draw [->] (pso1) -- node {\scriptsize{$\va_{3}^{h}$}} (c1n);
	\draw [->] (pson) -- node {\scriptsize{$\va_{n}^{h}$}} (cnn);
	
	\node (fitBox) [fill=lightgray!45,rounded corners,minimum 
	width=0.8cm,minimum height=4.02cm, anchor=north] at (3.45,1.8) {};
	
	\node(fit1) [fit, right=3mm of c1n] {\scriptsize{${J}_3$}};
	\node(fit2) [fit, right=3mm of c2n] {\scriptsize{${J}_2$}};
	\node(fit3) [fit, right=3mm of c3n] {\scriptsize{${J}_1$}};	
	\node(fitn) [fit, right=3mm of cnn] {\scriptsize{${J}_n$}};
	\path (fit1) -- node (dots4) [auto=false]{\vdots} (fitn); 
	
	\draw [->] (c2n) -- (fit2);
	\draw [->] (c3n) -- (fit3);
	\draw [->] (c1n) -- (fit1);
	\draw [->] (cnn) -- (fitn);
	
	\node (sort) [rectangle, text centered, draw=black, rounded corners, 
	fill=gray!30, right = 
	3mm of fitBox]{\scriptsize{Sort}};
	\draw[->](fitBox)--(sort);
	
	\node (dec1) [decision, right = 1.8cm of sort] {\scriptsize{$\ell_{i-1} - 
	\widehat{J}_{1} > \Delta_1$}}; 
	\draw[->] (sort) -- node{\scriptsize{$\{\widehat{J}_{k}\}_{k=1}^n$}}(dec1);
	\node (dec2) [decision,aspect=3, below = .5cm of dec1] {\scriptsize{${h} < 
	{h}_{max}$}};
	\node (dec4) [decision,aspect=3, below = 2.5cm of psoBox] 
	{\scriptsize{$i < m$}};
	\node (dec3) [decision,aspect=3, right = 2cm of dec4] 
	{\scriptsize{$\ell_i > \varphi$}};
	
	\node (incH) [rectangle, text centered, draw=black, rounded corners, 
	fill=gray!30, left = 5mm of dec2]{\scriptsize{$h\mathrel{++}$}};
	
	\node (dec5) [decision,aspect=3, below = 2mm of incH] {\scriptsize{${p} < 
	{p}_{max}$}};
	
	\node (incP) [rectangle, text centered, draw=black, text width=1.5cm, 
	rounded corners, fill=gray!30, left = 4mm of 
	dec5]{\scriptsize{$h:=1;$}\\\scriptsize{$p\mathrel{+}=p_{inc};$}};
	
	\node (incL) [rectangle, text centered, draw=black, rounded corners, 
	fill=gray!30, above = 5mm of dec4]{\scriptsize{$i\mathrel{++}$}};

	\node (stuff) [rectangle, text centered, draw=black, rounded corners, 
	fill=gray!30, right = 15mm of dec3]{\scriptsize{$\ell_i:=\widehat{J}_1$}};
    
	\node(sampleBox1) [fill=lightgray!45,rounded corners,minimum 
	width=1.4cm,minimum height=2.02cm, anchor=north, right = 6.87cm of c2n, text width = 1cm, align = left]{$\mathcal{I}$};
	\node(sampleBox2) [draw=black, rounded corners,minimum width=1.4cm,minimum 
	height=3.86cm, , right = 5.9cm of fitBox]{};
	\node [above = 1.5mm of sampleBox1,text width=2cm,align = 
	center]{\scriptsize{Resampling}};
	
	\node (c1r) [conf, right = 7.42cm of c3n] {};
	\node (c2r) [conf, right = 7.42cm of c1n]{};
	\path (c1r) -- node (dots4) [auto=false]{\vdots} (c2r);
	\node (c3r) [conf, cross, fill=TuInfRed, below = 2mm of c2r] {};
	\node (cnr) [conf, cross, fill=TuInfRed, right = 7.42cm of cnn] {};
	\path (c3r) -- node (dots4) [auto=false]{\vdots} (cnr);
	
	\node [output, right = 3mm of sampleBox1] (output) {};
	
	\draw[-] (sampleBox1) -- (output);
	\draw[->] (output) |- (c3r);
	\draw[->] (output) |- (cnr);
	
	\draw[->](dec1.east) -- node[near start](arr2){\scriptsize{Yes}} 
	(sampleBox2.west);
	\draw[->](dec1.south) -- node(arr3){\scriptsize{No}} (dec2.north);
	
	\draw[->](sampleBox2.south) |- node(arr1)[right,near 	
	start]{} (stuff.east);
	\draw[->](stuff.west) |- (dec3.east);
    
	\draw[->](dec3.west) -- node(arr1)[above]{\scriptsize{Yes}} (dec4.east);
	\draw[->] (dec4.north) -- node(arr5){\scriptsize{Yes}} (incL.south);
	\draw[->] (incL.north) -- (psoBox.south);
	\draw[->](dec2.west) -- node(arr6)[above,near start]{\scriptsize{Yes}} 
	(incH.east);
	\draw[->](incH.west) --  (incH -| incL);

	\draw[->](dec2.south) |- node[above,near end]{\scriptsize{No}} (dec5.east);
	\draw[->] (dec5.west) -- node(arr7)[above,near 
	start]{\scriptsize{Yes}}(incP);
	\draw[->] (incP) -- (incP -| incL);
	\node(V) [below = 4mm of dec3,text width=2cm,align = 
	center]{\scriptsize{\textbf{Stable state}}};
	\draw[->] (dec3.south) --node[right]{\scriptsize{No}} (V);
	\node(time) [below = 4mm of dec4,text width=2cm,align = 
	center]{\scriptsize{\textbf{Timeout}}};
	
	\draw[->] (dec4.south) --node[right]{\scriptsize{No}} (time);
	\node(exaust) [below = 4mm of dec5,text width=4cm,align = 
	center]{\scriptsize{\textbf{Particle exhaustion}}};
	\draw[->] (dec5.south) --node[right]{\scriptsize{No}} (exaust);
	
	\end{tikzpicture}
\vspace*{-2mm}
\caption{Graphical representation of ARES.}
	\label{fig:approach}
    \vspace*{-5mm}
\end{figure}
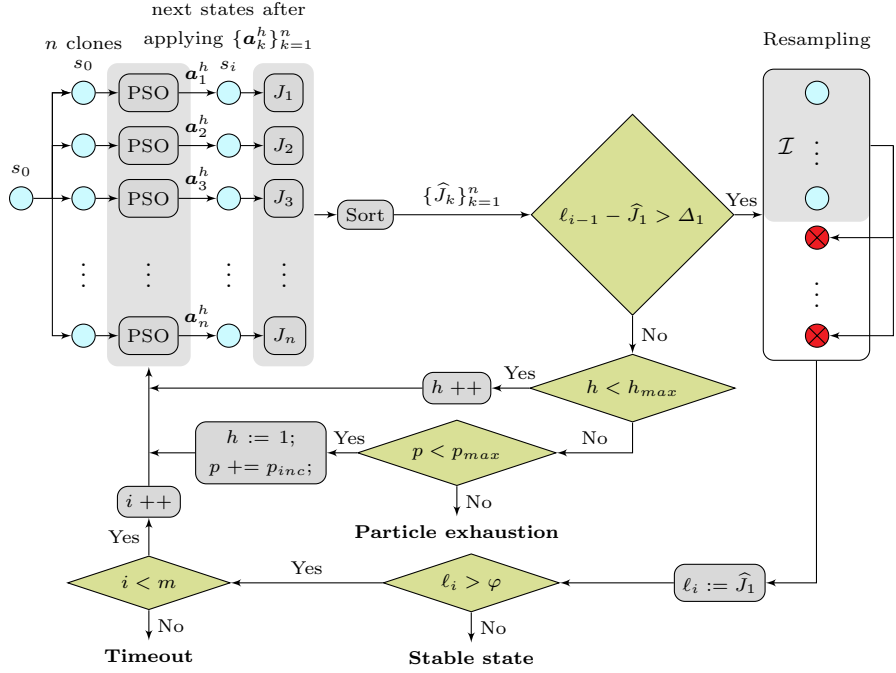
\begin{algorithm}[b]
	\SetKwFunction{Cost}{Cost}
	\SetKwFunction{ImportanceSplitting}{ImportanceSplitting}
	\SetKwFunction{particleswarm}{particleswarm}
    \SetKwFunction{Simulate}{Simulate}
	\SetKwInOut{Input}{Input}
	\SetKwInOut{Output}{Output}
\ForEach {$\mathcal{M}_k\in \mathcal{M}$}
{
    $[\va^{h}_k,\mathcal{M}^{h}_k] \leftarrow$ \particleswarm{$\mathcal{M}_k,{p},h$}; \textit{// use 
    PSO in 
    order to determine best next action for the MDP $\mathcal{M}_k$ with RPH $h$\\}
    ${J}_k(s_i)\leftarrow$ 
    \Cost{$\mathcal{M}_k^h,\va^{h}_k,{h}$};
    \textit{// calculate cost function if applying the sequence of optimal actions of length $h$}\\
    \If{${J}_k(s_{i-1})-{J}_k(s_i)>\Delta_k$}
    {
    $\Delta_k\leftarrow{J}_k(s_{i})/(m-i);$ \textit{// new level-threshold}\\
    }
}
	\caption{Simulate ($\mathcal{M},h,i,\{\Delta_k,{J}_k(s_{i-1})\}_{k=1}^n$)}
	\label{alg:sim}
\end{algorithm}
\setlength{\floatsep}{0.1cm}

The levels serve two purposes.  First, they implicitly define a Lyapunov function, which guarantees convergence.  If desired, this function can be explicitly generated for all states, up to some topological equivalence.  Second, the levels $\ell_{i}$ help PSO overcome local minima (see Fig.~\ref{fig:levels}~(left)).  If reaching a next level requires PSO to temporarily pass over a state-cost ridge, then ARES incrementally increases the size of the horizon $h$, up to a maximum size $h_{max}$. For particle $k$, passing the thresholds $\Delta_k$ means that it reaches a new level, and the definition of  $\Delta_k$ ensures a smooth degradation of its threshold.

Another idea imported from \gls{isp} and shown in Fig.~\ref{fig:approach}, is to maintain $n$ clones $\{\mathcal{M}_k\}_{k=1}^n$ of the MDP $\mathcal{M}$ (and its initial state) at any time $t$, and run \gls{pso}, for a horizon $h$, on each $h$-unfolding $\mathcal{M}^h_k$ of them. This results in an action sequence $\va^{h}_k$ of length $h$ (see Algo.~\ref{alg:sim}). This approach allows us to call  \gls{pso} for each clone and desired horizon, with a very small number of particles $p$ per clone.  

\SetAlgoSkip{}
\begin{algorithm}[t]
\BlankLine
	$\mathcal{I}\leftarrow$ Sort ascending $\mathcal{M}^h_{k}$ by their current costs;  \textit{// find indexes of MDPs whose costs are below the median among all the clones}\\
	\For{$k=1$ \KwTo $n$}
	{
		\eIf{$k\notin\mathcal{I}$}
		{
			Sample $r$ uniformly at random from $\mathcal{I}$;
			$\mathcal{M}_k \leftarrow \mathcal{M}_r^h$;\\
		}
		{
			$\mathcal{M}_k\leftarrow\mathcal{M}^h_k$; \textit{// Keep more successful MDPs unchanged} 
		}
	}
	\caption{Resample ($\{\mathcal{M}_k^h,{J}_k(s_{i})\}_{k=1}^n$)}
	\label{alg:resample}
\SetAlgoSkip{}
\end{algorithm}
\setlength{\floatsep}{0.1cm}

To check which particles have overcome their associated thresholds, we sort the particles according to their current cost, and split them in two sets: the successful set, having the indexes $\mathcal{I}$ and whose costs are lower than the median among all clones; and the unsuccessful set with indexes in $\{1,{\ldots},n\}\,{\setminus}\mathcal{I}$, which are discarded. The unsuccessful ones are further replenished, by sampling uniformly at random from the successful set $\mathcal{I}$ (see Algo.~\ref{alg:resample}).

The number of particles is increased $p\,{=}\,p\,{+}\,p_{inc}$ if no clone reaches a next level, for all horizons chosen.  Once this happens, we reset the horizon to one, and repeat the process. In this way, we adaptively focus our resources on escaping from local minima.  From the last level, we choose the state $s^{*}$ with the minimal cost, and traverse all of its predecessor states to find an optimal plan comprised of actions $\{\va^i\}_{1\leqslant i\leqslant m}$ that led MDP $\mathcal{M}$ to the optimal state $s^*$. In our running example, we select a flock in V-formation, and traverse all its predecessor flocks. The overall procedure of ARES is shown in Algo.~\ref{alg:ares}.
\SetAlgoSkip{}
\begin{algorithm}[!ht]
	\SetKwFunction{Fitness}{Fitness}
	\SetKwFunction{Resample}{Resample}
	\SetKwFunction{particleswarm}{particleswarm}
    \SetKwFunction{Simulate}{Simulate}
	\SetKwInOut{Input}{Input}
	\SetKwInOut{Output}{Output}
	
	\Input{$\mathcal{M},\varphi,{p}_{start},{p}_{inc},{p}_{max},{h}_{max},m,n$}
	\Output{$\{\va^i\}_{1\leqslant i\leqslant\,m}$ \textit{// synthesized optimal plans}}
	\BlankLine
	Initialize $\ell_0\leftarrow\inf$; $\{J_k(s_0)\}_{k=1}^n\leftarrow\inf$; ${p}\leftarrow{p}_{start}$; $i\leftarrow 1$;  ${h}\leftarrow 1$; $\Delta_k\leftarrow 0$;
	\BlankLine
		\While{($\ell_i > \varphi)$ $\vee$ $(i
	< m)$}
	{
     	\textit{// find and apply best actions with RPH $h$}\\
		$[\{\va_k^h,J_k(s_i),\mathcal{M}^h_{k}\}_{k=1}^n]\leftarrow$\Simulate{$\mathcal{M},h,i,\{\Delta_k,{J}_k(s_{i-1})\}_{k=1}^n$};
		$\widehat{J}_1 \leftarrow 
		sort({J}_1(s_i),\ldots,{J}_n(s_i))$; \textit{// find minimum cost among all the clones}\\	
		\eIf{$\ell_{i-1}-\widehat{J}_1>\Delta_1$}
		{
			$\ell_i\leftarrow\widehat{J}_1$; \textit{// new level has been reached}\\
            $i \leftarrow i + 1$;
			${h} \leftarrow 1$;
			${p} \leftarrow {p}_{start}$; \textit{// reset adaptive parameters}\\
			$\{\mathcal{M}_k\}_{k=1}^n\leftarrow$ 
			\Resample{$\{\mathcal{M}_k^h,{J}_k(s_{i})\}_{k=1}^n$};
		}
		{
			\eIf{${h} < {h}_{max}$}
			{
				${h} \leftarrow {h} + 1$; \textit{// improve time exploration}\\
			}
			{
				\eIf{${p} < {p}_{max}$}
				{
					${h} \leftarrow 1$;
					${p} \leftarrow {p} + 
					{p}_{inc}$; \textit{// improve space exploration}\\
				}
				{break;}
			}	
		}
	}
    Take a clone in the state with minimum cost $\ell_i=J(s^*_i)\leqslant\varphi$ at the last level $i$;\\
    \ForEach{$i$}
    {
    $\{s_{i-1}^*,\va^i\}\leftarrow Pre(s_i^*);$ \textit{// find predecessor and corresponding action}\\
    }
	\caption{ARES}
	\label{alg:ares}
\end{algorithm}
\setlength{\floatsep}{0.1cm}

%
\begin{proposition}[Optimality and Minimality] (1) Let $\mathcal{M}$ be an MDP.
For any initial state $s_0$ of $\mathcal{M}$, ARES is able to solve the optimal-plan synthesis problem for $\mathcal{M}$ and $s_0$. 
(2)~An optimal choice of $m$ in function $\Delta_k$, for some particle $k$, ensures that ARES also generates the shortest optimal plan.
\end{proposition}
\vspace*{-2mm}
\begin{proof}[Sketch]
(1)~The dynamic-threshold function $\Delta_k$ ensures that the initial cost in $s_0$ is continuously decreased until it falls below $\varphi$.  Moreover, for an appropriate number of clones, by adaptively determining the horizon and the number of particles needed to overcome $\Delta_k$, ARES always converges, with probability 1, to an optimal state, given enough time and memory. (2)~This follows from  convergence property (1), and from the fact that ARES always gives preference to the shortest horizon while trying to overcome $\Delta_k$.
\end{proof}
%
%
%

The optimality referred to in the title of the paper is in the sense of (1).  One, however, can do even better than (1), in the sense of (2), by empirically determining parameter $m$ in the dynamic-threshold function $\Delta_k$.  Also note that ARES is an \emph{approximation algorithm}. As a consequence, it might return nonminimal plans.  Even in these circumstances, however, the plans will still lead to an optimal state.  This is a V-formation in our flocking example.

\newcommand{\majExp}{{8,000}}

\section{Experimental Results}
\label{sec:results}
To assess the performance of our approach, we developed a simple simulation environment in Matlab. All experiments were run on an Intel Core i7-5820K CPU with 3.30 GHz and with 32GB RAM available. 

We performed numerous experiments with a varying number of birds. Unless stated otherwise, results refer to {\majExp} experiments with 7 birds with the following parameters: ${p}_{start}\,{=}\,10$, ${p}_{inc}\,{=}\,5$, ${p}_{max}\,{=}\,40$, $\ell_{max}\,{=}\,20$, ${h}_{max}\,{=}\,5$, $\varphi\,{=}\,10^{-3}$, and $n\,{=}\,20$. The initial configurations were generated independently uniformly at random subject to the following constraints: 
\begin{enumerate}\itemsep=0em
\item Position constraints: $\forall\:i\,{\in}\,\{1,{\ldots},7\}.\:\xv_i(0)\in[0,3]\times[0,3]$.
\vspace*{1mm}\item Velocity constraints: $\forall\:i\,{\in}\,\{1,{\ldots},7\}.\:\vv_i(0)\in[0.25,0.75]\times[0.25,0.75]$.
\end{enumerate}

\begin{figure}[t]
	\centering
	\includegraphics[width=.49\textwidth]{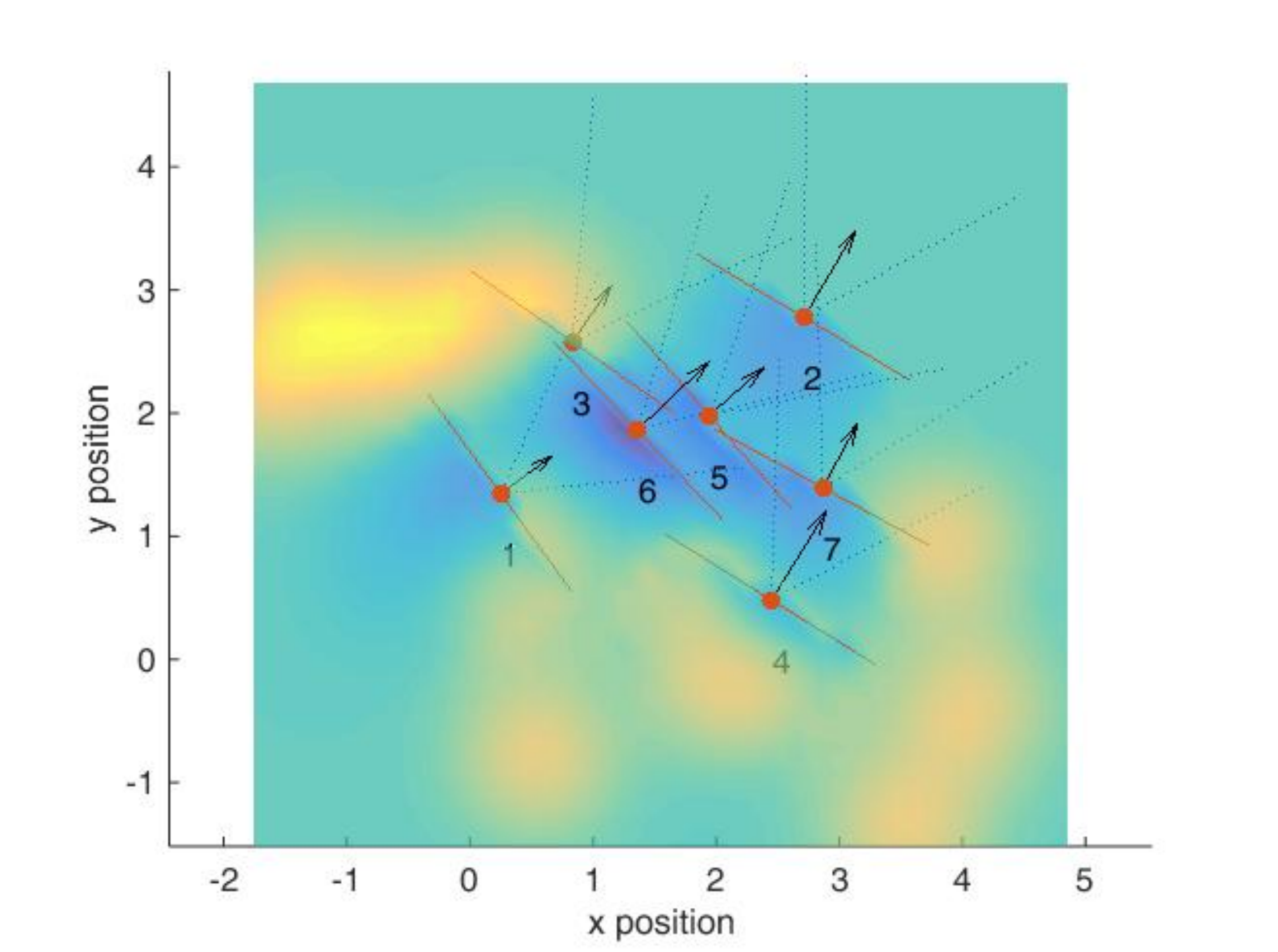}
	\includegraphics[width=.49\textwidth]{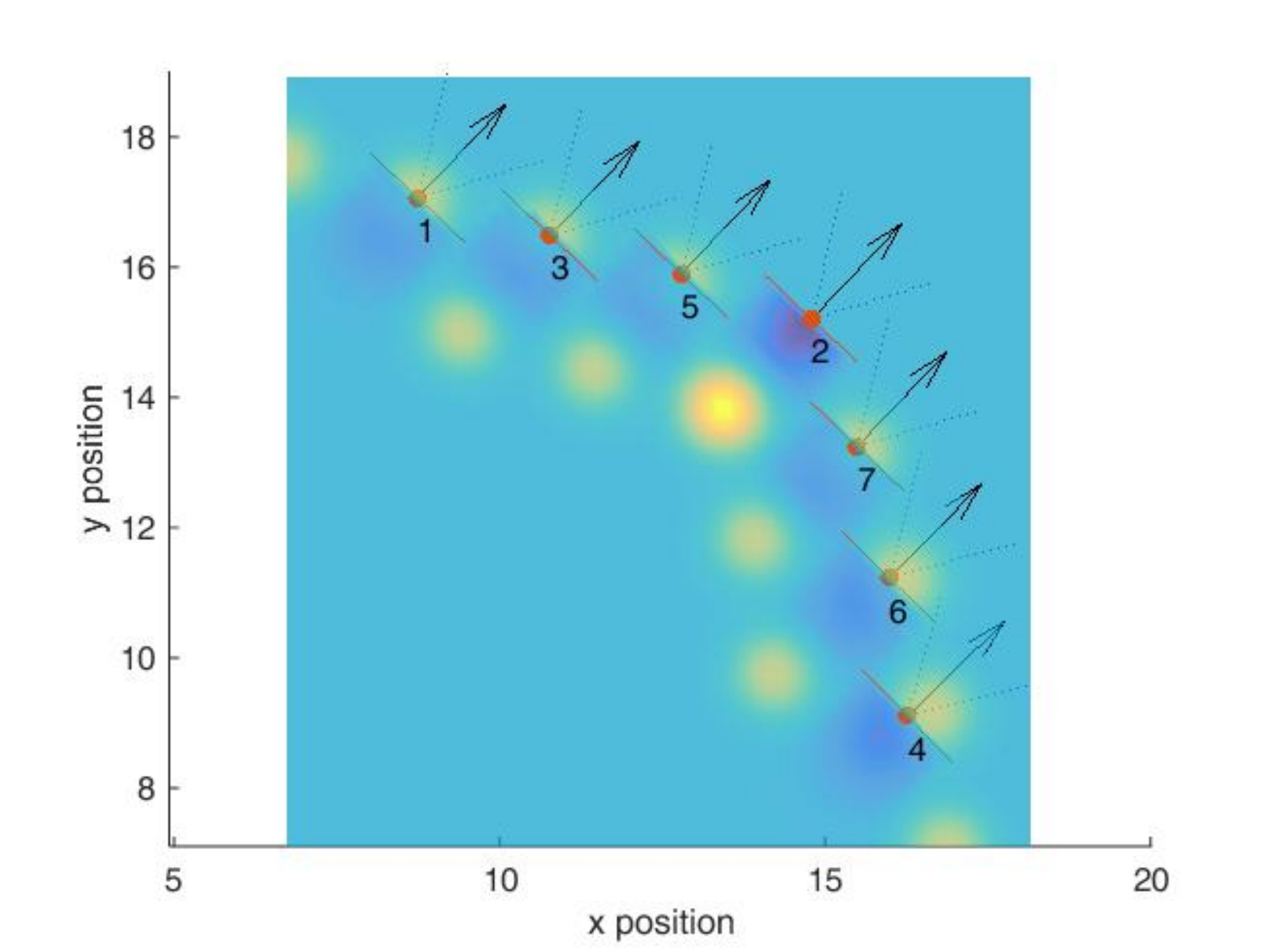}
    \vspace*{-2mm}
\caption{Left: Example of an arbitrary initial configuration of 7 birds. Right: The V-formation obtained by applying the plan generated by ARES. In the figures, we show the wings of the birds, bird orientations,  bird speeds (as scaled arrows), upwash regions in yellow, and downwash regions in dark blue.}
    \label{fig:form}
    \vspace*{3mm}
\end{figure}

\begin{table}
	\scriptsize
	\centering
	\caption{Overview of the results for \majExp$\:$experiments with 7 birds}
	\begin{tabular}{l c c c c c c c c }  
		\toprule
		& \multicolumn{4}{c}{{\textsc{Successful}}} & 
		\multicolumn{4}{c}{{\textsc{Total}}} \\
		\cmidrule(l){2-5}\cmidrule(l){6-9}
		No. Experiments~~~~ & \multicolumn{4}{c}{7573}  & 
		\multicolumn{4}{c}{8000} 
		\\
		\cmidrule(l){2-5}\cmidrule(l){6-9}
		& {\centering\textsc{Min}} & {\textsc{Max}}& {\centering\textsc{Avg}} & 
		{\textsc{Std}} 
		& {\textsc{Min}} & {\textsc{Max}} & {\textsc{Avg}} & 
		{\textsc{Std}} \\
		
		\midrule 
		Cost, ${J}$
			& 2.88$\cdot10^{-7}$	& 9$\cdot10^{-4}$ 	&	4$\cdot10^{-4}$  & 3$\cdot10^{-4}$	
			& 2.88$\cdot10^{-7}$	& 1.4840 	&	0.0282  & 0.1607   	\\
		Time, $t$
			& 23.14s 	& 310.83s 	& 63.55s	& 22.81s	
			& 23.14s  	& 661.46s 	& 64.85s 	& 28.05s	\\
		Plan Length, $i$		
			& 7 		& 20 		& 12.80		& 2.39 			
			& 7 		& 20 		& 13.13		& 2.71  \\
		RPH, ${h}$	
			& 1 		& 5 		& 1.40		& 0.15 					
			& 1 		& 5 		& 1.27		& 0.17 \\
	\bottomrule
\end{tabular}
\label{tab:res5000overview}
\vspace*{-3mm}
\end{table}

\begin{table}
\vspace*{-3mm}
	\scriptsize
	\centering
	\caption{Average duration for 100 experiments with various number of birds}
	\begin{tabular}{l cc c c c }  
		\toprule
		No. of birds & & 3 & 5 & 7 & 9 \\
      	\cmidrule(l){2-6}
        Avg. duration & & 4.58s & 18.92s & 64.85s & 269.33s \\
		\bottomrule
\end{tabular}
\label{tab:res100time}
\end{table}

Table~\ref{tab:res5000overview} gives an overview of the results with respect to the \majExp$\:$ experiments we performed with 7 birds for a maximum of 20 levels. The average fitness across all experiments is at $0.0282$ with a standard deviation of $0.1654$. We achieved a success rate of $94.66\%$ with fitness threshold $\varphi=10^{-3}$. The average fitness is higher than the threshold due to comparably high fitness of unsuccessful experiments. When increasing the bound for the maximal plan length $m$ to 30 we achieved a $98.4\%$ success rate in 1,000 experiments at the expense of a slightly longer average execution time.

\begin{figure}[!htb]
	\centering
    \includegraphics[width=.328\textwidth]{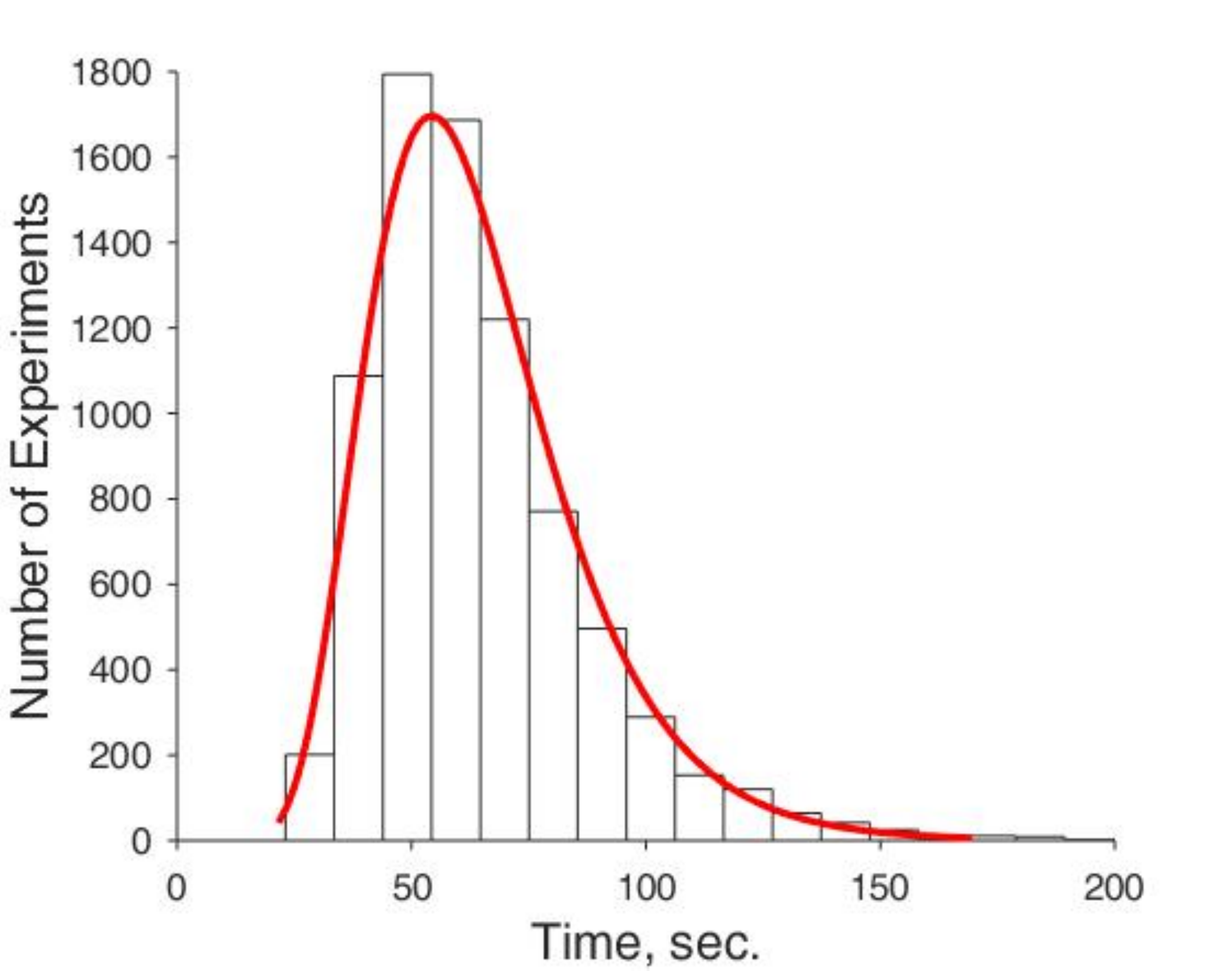}
	\includegraphics[width=.328\textwidth]{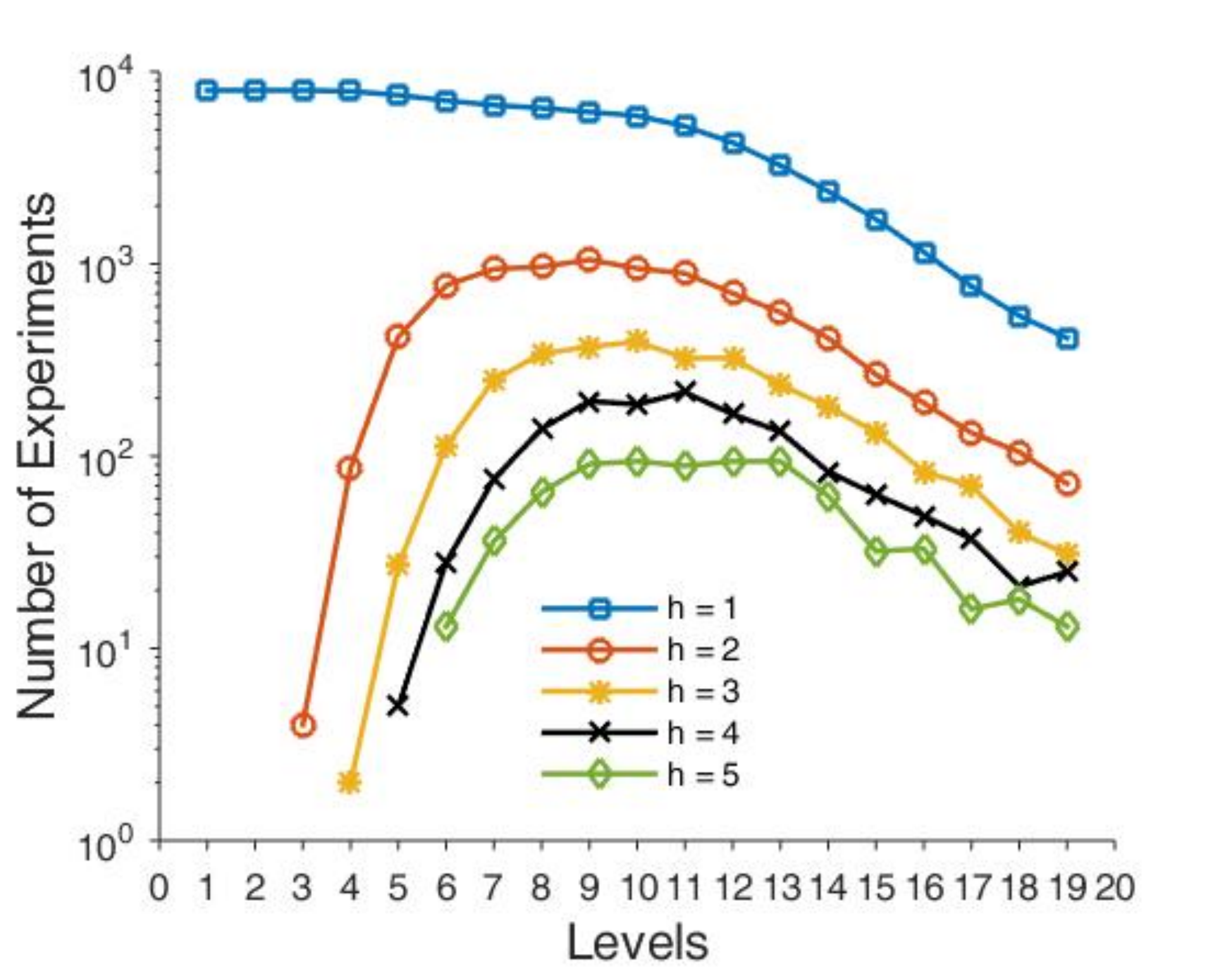}
	\includegraphics[width=.328\textwidth]{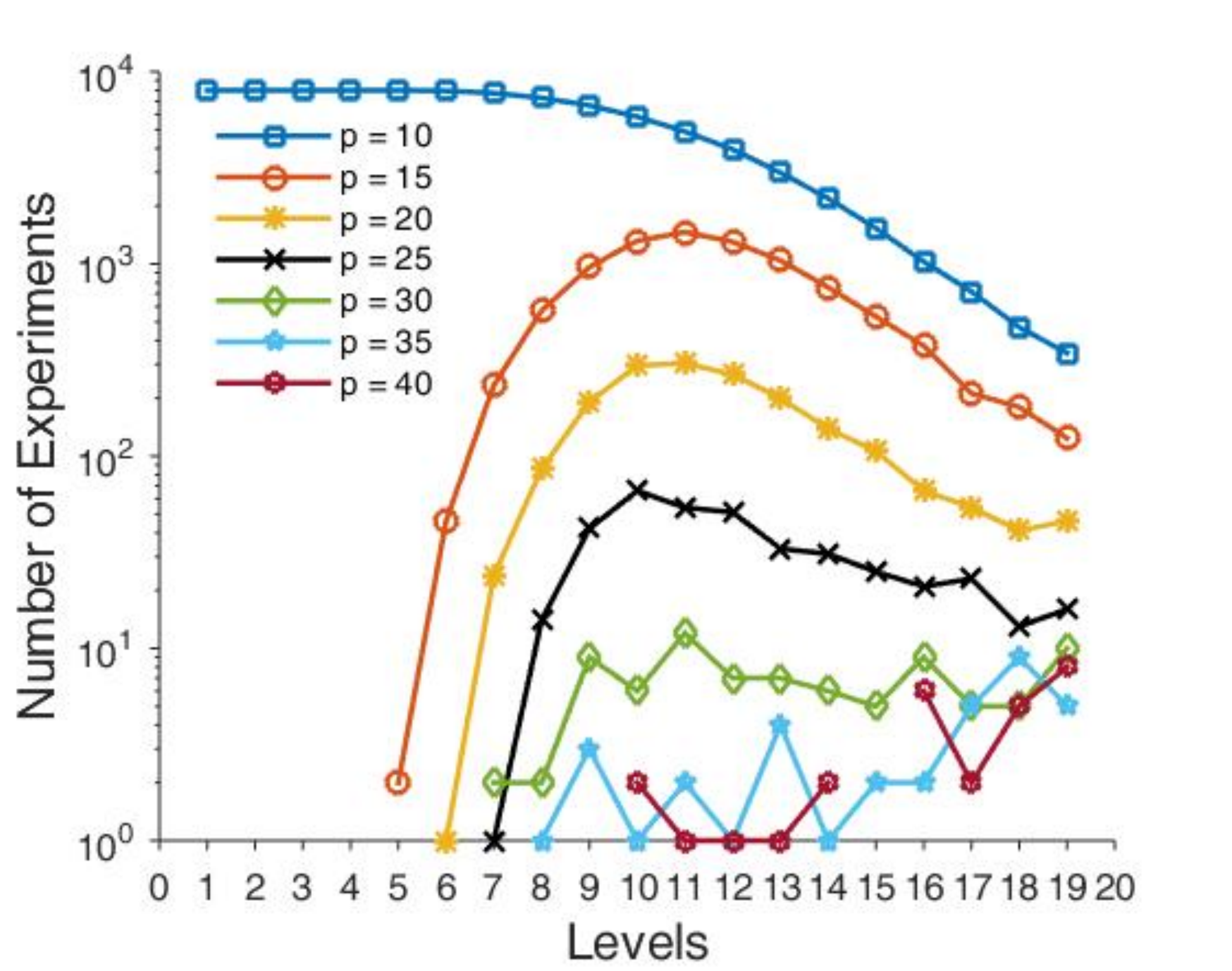}
	\caption{Left: Distribution of execution times for \majExp$\:$runs. Middle: Statistics of increasing RPH $h$. 
Right: Particles of PSO ${p}$ for \majExp$\:$experiments}\label{fig:aheads_pso}
   \vspace*{-3mm}
\end{figure}

The left plot in Fig.~\ref{fig:aheads_pso} depicts the resulting distribution of execution times for \majExp$\:$ runs of our algorithm, where it is clear that, excluding only a few outliers from the histogram, an arbitrary configuration of birds (Fig.~\ref{fig:form} (left)) reaches V-formation  (Fig.~\ref{fig:form} (right)) in around 1~minute. The execution time rises with the number of birds as shown in Table~\ref{tab:res100time}.

In Fig.~\ref{fig:aheads_pso}, we illustrate for how many experiments the algorithm had to increase RPH $h$ (Fig.~\ref{fig:aheads_pso} (middle)) and the number of particles used by PSO ${p}$ (Fig.~\ref{fig:aheads_pso} (right))  to improve time and space exploration, respectively.

After achieving such a high success rate of ARES for an arbitrary initial configuration, we would like to demonstrate that the number of experiments performed is sufficient for high confidence in our results. This requires us to determine the appropriate number $N$ of random variables $Z_1, ... Z_N$ necessary for the Monte-Carlo approximation scheme we apply to assess efficiency of our approach. For this purpose, we use the additive approximation algorithm as discussed in~\cite{grosu2014isola}. If the sample mean $\mu_Z\,{=}\,(Z_1\,{+}\,{\ldots}\,{+}\,Z_N)/N$ is expected to be large, then one can exploit the Bernstein's inequality and fix $N$ to $\Upsilon\,{\propto}\,ln(1/\delta)/\varepsilon^2$. This results in an {\em additive} or {\em absolute-error $(\varepsilon,\delta)$-approximation scheme}:
\[
{\bf P}[\mu_Z\,{-}\,\varepsilon\leq\widetilde{\mu}_Z\leq\mu_Z\,{+}\,\varepsilon)]\geq{}1-\delta,
\]
where $\widetilde{\mu}_Z$ approximates $\mu_Z$ with absolute error $\varepsilon$ and probability $1-\delta$.

In particular, we are interested in $Z$ being a Bernoulli random variable:
 
 \[Z=\left\{
\begin{array}{ll}
1, & \text{if}\:\:J(\boldsymbol{c}(t),\va(t),{h}(t))\leqslant\varphi,\\
0, & \text{otherwise}.
\end{array}\right.\]

Therefore, we can use the Chernoff-Hoeffding instantiation of the Bernstein's inequality, and further fix the proportionality constant to $\Upsilon\,{=}\,4\,ln(2/\delta)/\varepsilon^2$, as in~\cite{HLMP04}. 
Hence, for our performed \majExp$\:$experiments, we achieve a success rate of 95\% with absolute error of $\varepsilon = 0.05$ and confidence ratio 0.99.

Moreover, considering that the average length of a plan is 13, and that each state in a plan is independent from all other plans, we can roughly consider that our above estimation generated 80,000 independent states. For the same confidence ratio of 0.99 we then obtain an approximation error $\varepsilon\,{=}\,0.016$, and for a confidence ratio of 0.999, we obtain an approximation error $\varepsilon\,{=}\,0.019$.




	\section{Related Work}
\label{sec:related}

Organized flight in flocks of birds can be categorized in \emph{cluster 
flocking} and \emph{line formation}~\cite{heppner1974avian}. In cluster 
flocking the individual birds 
in a large flock seem to be uncoordinated in general. However, the flock moves, 
turns, and wheels as if it were one organism. In 1987 
Reynolds~\cite{Reynolds1987CG} defined his three famous rules describing 
separation, alignment, and cohesion for individual birds in order to have them 
flock together. This work has been great inspiration for research in the area 
of collective behavior and self-organization.

In contrast, line formation flight requires the individual birds to fly in a 
very specific formation. Line formation has two main benefits for the 
long-distance migrating birds. First, exploiting the generated uplift by birds 
flying in front, trailing birds are able to conserve 
energy~\cite{lissaman1970formation,Cutts251,weimerskirch2001nature}. Second, in 
a staggered formation, all birds have a clear view in front as well as a view 
on their neighbors~\cite{Bajec2009AB}. While there has been quite some effort 
to keep a certain formation for multiple entities when traveling 
together~\cite{Seiler2002CDC, Gennaro2005CAIC,Dang2015CYBCONF}, only little 
work deals with a task of achieving this extremely important formation from a 
random starting configuration \cite{Cattivelli2011TSP}. The convergence of bird 
flocking into V-formation has been also analyzed with the use of combinatorial 
techniques\cite{Chazelle:2014}. 


Compared to previous work, in~\cite{MPC2007} this question is addressed without using any behavioral rules but as problem of \emph{optimal control}. In~\cite{yang2016love} a cost function was proposed that reflects all major features of V-formation, namely, \emph{Clear View} (CV), \emph{Velocity Matching} (VM), and \emph{Upwash Benefit} (UB). The technique of \gls{mpc} is used to achieve V-formation starting from an arbitrary initial configuration of $n$ birds. \gls{mpc} solves the task by minimizing a functional defined as squared distance from the optimal values of CV, VM, and UB, subject to constraints on input and output. The approach is to choose an optimal \emph{velocity adjustment}, as a control input, at each time-step applied to the velocity of each bird by predicting model behavior several time-steps ahead. 
 
The controller synthesis problem has been widely studied~\cite{Plans2012}.
The most popular and natural technique is Dynamic Programming (DP)~\cite{Bellman:1957} that improves the approximation of the functional at each iteration, eventually converging to the optimal one given a fixed asymptotic error. Compared to DP, which considers all the possible states of the system and might suffer from state-space explosion in case of environmental uncertainties, approximate algorithms~\cite{Henriques2012,Bartocci2016,mannor_cross_2003,bartlett_experiments_2011,stulp_policy_2012,stulp_path_2012} take into account only the paths leading to desired target. One of the most efficient ones is Particle Swarm Optimization (PSO)~\cite{Kennedy95particleswarm} that has been adopted for finding the next best step of MPC in~\cite{yang2016love}. Although it is a very powerful optimization technique, it has not yet been possible to achieve a high success rate in solving the considered flocking problem. Sequential Monte-Carlo methods proved to be efficient in tackling the question of control for linear stochastic systems~\cite{chen2009fast}, in particular, Importance Splitting (IS)~\cite{KalajdzicIsola16}. The approach we propose is, however, the first attempt to combine adaptive IS, PSO, and receding-horizon technique for \emph{synthesis of optimal plans for controllable systems}. 
We use MPC to synthesize a plan, but use IS to determine the intermediate fitness-based waypoints.  We use PSO to solve the multi-step optimization problem generated by MPC,  but choose the planning horizon and the number of particles adaptively. These choices are governed by the difficulty to reach the next level.

	\section{Conclusion and Future Work}
\label{sec:conclusion}

In this paper, we have presented ARES, a very general adaptive, receding-horizon synthesis algorithm for MDP-based optimal plans.  Additionally, ARES can be readily converted into a
model-predictive controller with an adaptive receding horizon and statistical guarantees of convergence.  We have also conducted a very thorough performance analysis of ARES based on the problem of V-formation in a flock of birds.  For flocks of 7
birds, ARES is able to generate an optimal plan leading to a
V-formation in 95\% of the 8,000 random initial configurations we considered, with an average
execution time of only 63 seconds per plan.

The execution time of the ARES algorithm can be even further improved in a number of ways.
First, we currently do not parallelize 
our implementation of the PSO algorithm.  Recent work
\cite{Hung2012,Rymut2013,Zhou2009} has shown how Graphic Processing Units 
(GPUs) are very efficient at accelerating PSO computation.
Modern GPUs, by providing thousands of cores, are well-suited for implementing PSO as they enable execution of a very large number of particles in parallel, which can improve accuracy of the optimization procedure.
Likewise, the calculation of the fitness function can also be 
run in parallel.  The parallelization of these steps 
should significantly speed up our simulations.

Second, we are currently using a static approach to decide how to increase our 
prediction horizon and the number of particles used in PSO. Specifically, we first 
increase the prediction horizon from 1 to 5, while 
keeping the number of particles unchanged at 10; if this fails  to find a solution with fitness $\widehat{J_1}$ satisfying 
$\ell_{i-1}\,{-}\,\widehat{J_1}>\Delta_1$, we then increase the 
number of particles by 5. Based on our results, we speculate 
that in the initial stages, increasing the prediction horizon is more beneficial (leading rapidly to
the appearance of cost-effective formations), whereas in the later stages, increasing the number 
of particles is more helpful.
As future work, we will use machine-learning approaches to decide on the prediction horizon and the number of particles deployed at runtime given the current level and state of the MDP.

Third, in our approach, we always calculate the number of clones for 
resampling based on the current state. An alternative approach would rely on 
statistics built up over multiple levels in combination with the rank in the 
sorted list to determine whether a configuration should be used for 
resampling or not.

Finally, we are currently using our approach to generate plans for a flock to go
from an initial configuration to a final V-formation. Our eventual
goal is to achieve formation flight for a robotic swarm of (bird-like) drones.
A real-world example is parcel-delivering drones that follow the same route to their destinations.  
Letting them fly together for a while could save energy and increase flight time.
To achieve this goal, we first need to investigate the wind dynamics of  multi-rotor drones.
Then, the fitness function needs to be adopted to the new wind dynamics.
Lastly, a decentralized approach of this method needs to be implemented and tested on the drone firmware.

\vspace*{3mm}\noindent{}{\bf Acknowledgments.}
%
%
The first author and the last author would like to thank Jan K\u{r}et\'{i}nsk\'{y} for very valuable feedback. This work was partially supported by the Doctoral Program
Logical Methods in Computer Science funded by the Austrian
Science Fund (FWF) project W1255-N23, and the Austrian National Research Network (nr. S
11405-N23 and S 11412-N23) SHiNE funded by FWF.

%

	\bibliographystyle{splncs03}
	\bibliography{flockingB}

\begin{thebibliography}{10}
\providecommand{\url}[1]{\texttt{#1}}
\providecommand{\urlprefix}{URL }

\bibitem{Bajec2009AB}
Bajec, I.L., Heppner, F.H.: Organized flight in birds. {Animal Behaviour}
  78(4),  777--789 (2009)

\bibitem{Bartocci2016}
Bartocci, E., Bortolussi, L., Br{\'a}zdil, T., Milios, D., Sanguinetti, G.:
  Policy learning for time-bounded reachability in continuous-time markov
  decision processes via doubly-stochastic gradient ascent. In: Proc. of QEST
  2016: the 13th International Conference on Quantitative Evaluation of
  Systems. vol. 9826, pp. 244--259 (2016)

\bibitem{bartlett_experiments_2011}
Baxter, J., Bartlett, P.L., Weaver, L.: Experiments with infinite-horizon,
  policy-gradient estimation. J. Artif. Int. Res.  15(1),  351--381 (2011)

\bibitem{Bellman:1957}
Bellman, R.: Dynamic Programming. Princeton University Press (1957)

\bibitem{MPC2007}
Camacho, E.F., Alba, C.B.: {Model Predictive Control}. {Advanced Textbooks in
  Control and Signal Processing}, Springer (2007)

\bibitem{Cattivelli2011TSP}
Cattivelli, F.S., Sayed, A.H.: Modeling bird flight formations using diffusion
  adaptation. IEEE Transactions on Signal Processing  59(5),  2038--2051 (2011)

\bibitem{CerouGuyader2007}
C\'erou, F., Guyader, A.: Adaptive multilevel splitting for rare event
  analysis. Stochastic Analysis and Applications  25,  417--443 (2007)

\bibitem{Chazelle:2014}
Chazelle, B.: {The Convergence of Bird Flocking}. Journal of the ACM  61(4),
  21:1--21:35 (2014)

\bibitem{chen2009fast}
Chen, Y., Wu, B., Lai, T.L.: {Fast Particle Filters and Their Applications to
  Adaptive Control in Change-Point ARX Models and Robotics}. INTECH Open Access
  Publisher (2009)

\bibitem{Cutts251}
Cutts, C., Speakman, J.: Energy savings in formation flight of pink-footed
  geese. Journal of Experimental Biology  189(1),  251--261 (1994)

\bibitem{Dang2015CYBCONF}
Dang, A.D., Horn, J.: Formation control of autonomous robots following desired
  formation during tracking a moving target. In: Proceedings of the
  International Conference on Cybernetics. pp. 160--165. IEEE (2015)

\bibitem{dimock2003aerodynamic}
Dimock, G., Selig, M.: {The Aerodynamic Benefits of Self-Organization in Bird
  Flocks}. Urbana  51,  1--9 (2003)

\bibitem{flake1998computational}
Flake, G.W.: {The Computational Beauty of Nature: Computer Explorations of
  Fractals, Chaos, Complex Systems, and Adaptation}. MIT Press (1998)

\bibitem{mpc1989}
García, C.E., Prett, D.M., Morari, M.: Model predictive control: Theory and
  practice -- a survey. Automatica  25(3),  335--348 (1989)

\bibitem{Gennaro2005CAIC}
Gennaro, M.C.D., Iannelli, L., Vasca, F.: {Formation Control and Collision
  Avoidance in Mobile Agent Systems}. In: Proceedings of the International
  Symposium on Control and Automation Intelligent Control. pp. 796--801. IEEE
  (2005)

\bibitem{Glasserman1999}
Glasserman, P., Heidelberger, P., Shahabuddin, P., Zajic, T.: Multilevel
  {S}plitting for {E}stimating {R}are {E}vent {P}robabilities. Operations
  Research  47(4),  585--600 (1999)

\bibitem{grosu2014isola}
Grosu, R., Peled, D., Ramakrishnan, C.R., Smolka, S.A., Stoller, S.D., Yang,
  J.: Using statistical model checking for measuring systems. In: {Proceedings
  of the International Symposium Leveraging Applications of Formal Methods,
  Verification and Validation}. LNCS, vol. 8803, pp. 223--238. Springer (2014)

\bibitem{Henriques2012}
Henriques, D., Martins, J.G., Zuliani, P., Platzer, A., Clarke, E.M.:
  Statistical model checking for markov decision processes. In: Proc. of QEST
  2012: the Ninth International Conference on Quantitative Evaluation of
  Systems. pp. 84--93. QEST'12, IEEE Computer Society (2012)

\bibitem{heppner1974avian}
Heppner, F.H.: Avian flight formations. Bird-Banding  45(2),  160--169 (1974)

\bibitem{HLMP04}
H\'{e}rault, T., Lassaigne, R., Magniette, F., Peyronnet, S.: Approximate
  probabilistic model checking. In: Proceedings of the International Conference
  on Verification, Model Checking, and Abstract Interpretation (2004)

\bibitem{Hung2012}
Hung, Y., Wang, W.: Accelerating parallel particle swarm optimization via gpu.
  Optimization Methods and Software  27(1),  33--51 (2012)

\bibitem{Kennedy95particleswarm}
James, K., Russell, E.: Particle swarm optimization. In: Proceedings of 1995
  IEEE International Conference on Neural Networks. pp. 1942--1948 (1995)

\bibitem{KalajdzicIsola16}
Kalajdzic, K., J{\'{e}}gourel, C., Lukina, A., Bartocci, E., Legay, A., Smolka,
  S.A., Grosu, R.: {Feedback Control for Statistical Model Checking of
  Cyber-Physical Systems}. In: Proceedings of the International Symposium
  Leveraging Applications of Formal Methods, Verification and Validation:
  Foundational Techniques. pp. 46--61. LNCS, Springer (2016)

\bibitem{lissaman1970formation}
Lissaman, P., Shollenberger, C.A.: Formation flight of birds. Science
  168(3934),  1003--1005 (1970)

\bibitem{mannor_cross_2003}
Mannor, S., Rubinstein, R.Y., Gat, Y.: The cross entropy method for fast policy
  search. In: {ICML}. pp. 512--519 (2003)

\bibitem{nathan2008}
Nathan, A., Barbosa, V.C.: {V-like Formations in Flocks of Artificial Birds}.
  Artificial Life  14(2),  179--188 (2008)

\bibitem{Reynolds1987CG}
Reynolds, C.W.: Flocks, herds and schools: A distributed behavioral model.
  SIGGRAPH Computer Graphics  21(4),  25--34 (1987)

\bibitem{russellnorvig}
Russell, S., Norvig, P.: Artificial Intelligence: {A} Modern Approach.
  Prentice-Hall, 3rd edn. (2010)

\bibitem{Rymut2013}
Rymut, B., Kwolek, B., Krzeszowski, T.: {GPU-Accelerated Human Motion Tracking
  Using Particle Filter Combined with PSO}. In: Proceedings. of the
  International Conference on Advanced Concepts for Intelligent Vision Systems.
  LNCS, vol. 8192, pp. 426--437. Springer (2013)

\bibitem{Seiler2002CDC}
Seiler, P., Pant, A., Hedrick, K.: Analysis of bird formations. In: Proceedings
  of the Conference on Decision and Control. vol.~1, pp. 118--123 vol.1. IEEE
  (2002)

\bibitem{stulp_path_2012}
Stulp, F., Sigaud, O.: Path integral policy improvement with covariance matrix
  adaptation. arXiv preprint arXiv:1206.4621  (2012),
  \url{http://arxiv.org/abs/1206.4621}

\bibitem{stulp_policy_2012}
Stulp, F., Sigaud, O.: Policy improvement methods: {Between} black-box
  optimization and episodic reinforcement learning (2012),
  \url{http://hal.upmc.fr/hal-00738463/}

\bibitem{Plans2012}
Verfaillie, G., Pralet, C., Teichteil, F., Infantes, G., Lesire, C.: {Synthesis
  of plans or policies for controlling dynamic systems}. {AerospaceLab} (4),
  p. 1--12 (2012)

\bibitem{weimerskirch2001nature}
Weimerskirch, H., Martin, J., Clerquin, Y., Alexandre, P., Jiraskova, S.:
  {Energy Saving in Flight Formation}. Nature  413(6857),  697--698 (2001)

\bibitem{yang2016love}
Yang, J., Grosu, R., Smolka, S.A., Tiwari, A.: {Love Thy Neighbor: V-Formation
  as a Problem of Model Predictive Control}. In: LIPIcs-Leibniz International
  Proceedings in Informatics. vol.~59. Schloss Dagstuhl-Leibniz-Zentrum fuer
  Informatik (2016)

\bibitem{yang2016bda}
Yang, J., Grosu, R., Smolka, S.A., Tiwari, A.: {V-Formation as Optimal
  Control}. In: Proceedings of the Biological Distributed Algorithms Workshop
  2016 (2016)

\bibitem{Zhou2009}
Zhou, Y., Tan, Y.: {GPU-based Parallel Particle Swarm Optimization}. In:
  Proceedings of the Congress on Evolutionary Computation. pp. 1493--1500. IEEE
  (2009)

\end{thebibliography}


\end{document}